\documentclass{article}

\usepackage{microtype}
\usepackage{graphicx}
\usepackage{subcaption}
\usepackage{booktabs} 
\usepackage{multirow} 

\usepackage{hyperref}

\usepackage[preprint]{icml2026}

\usepackage{amsmath}
\usepackage{amssymb}
\usepackage{mathtools}
\usepackage{amsthm}

\usepackage[capitalize,noabbrev]{cleveref}

\theoremstyle{plain}

\theoremstyle{definition}

\theoremstyle{remark}

\usepackage[textsize=tiny]{todonotes}

\icmltitlerunning{Submission and Formatting Instructions for ICML 2026}

\begin{document}

\twocolumn[
  \icmltitle{AutoQRA: Joint Optimization of Mixed-Precision Quantization and Low-rank Adapters for Efficient LLM Fine-Tuning}



  \icmlsetsymbol{equal}{*}

  \begin{icmlauthorlist}
    \icmlauthor{Changhai Zhou}{1}
    \icmlauthor{Shiyang Zhang}{2}
    \icmlauthor{Yuhua Zhou}{3}
    \icmlauthor{Qian Qiao}{4}
    \icmlauthor{Jun Gao}{2}
    \icmlauthor{Cheng Jin}{1}
    \icmlauthor{Kaizhou Qin}{1}
    \icmlauthor{Weizhong Zhang}{1}
  \end{icmlauthorlist}

  \icmlaffiliation{1}{Fudan University}
  \icmlaffiliation{2}{Yale University}
  \icmlaffiliation{3}{Zhejiang University}
  \icmlaffiliation{4}{OpenWPLab}
\icmlcorrespondingauthor{Changhai Zhou}{zhouch23@m.fudan.edu.cn}
  \icmlcorrespondingauthor{Cheng Jin}{jc@fudan.edu.cn}
  \icmlcorrespondingauthor{Weizhong Zhang}{weizhongzhang@fudan.edu.cn}

  \icmlkeywords{Machine Learning, ICML}

  \vskip 0.3in
]



\printAffiliationsAndNotice{}  

\begin{abstract}
Quantization followed by parameter-efficient fine-tuning has emerged as a promising paradigm for downstream adaptation under tight GPU memory constraints. However, this sequential pipeline fails to leverage the intricate interaction between quantization bit-width and LoRA rank. Specifically, a carefully optimized quantization allocation with low quantization error does not always translate to strong fine-tuning performance, and  different bit-width and rank configurations can lead to significantly varying outcomes under the same memory budget. To address this limitation, we propose \textbf{AutoQRA}, a joint optimization framework that simultaneously optimizes the bit-width and LoRA rank configuration for each layer during the mixed quantized fine-tuning process. To tackle the challenges posed by the large discrete search space and the high evaluation cost associated with frequent fine-tuning iterations, AutoQRA decomposes the optimization process into two stages. First, it first conducts a global multi-fidelity evolutionary search, where the initial population is warm-started by injecting layer-wise importance priors. 
This stage employs specific operators and a performance model to efficiently screen candidate configurations. Second, trust-region Bayesian optimization is applied to locally refine promising regions of the search space and identify optimal configurations under the given memory budget. This approach enables active compensation for quantization noise in specific layers during training. Experiments show that AutoQRA achieves performance close to full-precision fine-tuning with a memory footprint comparable to uniform 4-bit methods.
\end{abstract}    
\section{Introduction}\label{sec:intro}
Deploying large language model (LLMs) for specific downstream tasks can be prohibitively memory intensive, which prevents many users from adapting strong base models in practice \citep{makridakis2023large,raiaan2024review,chang2024survey}. A common workaround is a sequential pipeline: first quantize the pretrained backbone to fit a tight GPU memory budget, then perform parameter efficient fine tuning by training lightweight adapters such as LoRA while keeping the quantized backbone frozen \citep{hu2022lora,li2023loftq,QA-LoRA}. In this setting, the deployment budget is a hard constraint, and the objective is the post fine tuning performance achieved within that budget.

Recent work has begun to exploit layerwise heterogeneity in the quantize-then-fine-tune pipeline for LLMs. On the quantization side, mixed-precision methods allocate different bit widths across layers according to estimated sensitivity, aiming to reduce the discrepancy between the quantized model and its full-precision counterpart \citep{huang2025slimllm,amq}. On the adaptation side, non-uniform rank allocation concentrates LoRA capacity on layers that are more important for task adaptation \citep{zhang2023adalora,zhou2025rankadaptor}. However, we find that a bit-width allocation that appears favorable under reconstruction or calibration criteria can still lead to poor performance after fine-tuning. Moreover, under the same memory budget, different combinations of bit width and LoRA rank can yield sharply different outcomes. The fundamental reason is that bit width and LoRA rank interact, yet most methods treat them as independent decisions. For example, a typical performance-oriented pipeline first fixes per-layer precision using static proxies and then adjusts non-uniform ranks. This separation is misaligned with the deployment objective because the two knobs are coupled during training. Lower precision introduces quantization noise, while additional adapter capacity can partially compensate for that noise through learning. Once bit widths are fixed, the system loses the opportunity to trade redundant precision for learnability in the layers where adapters can use it most effectively, leading to systematic resource misallocation.

These observations motivate a joint optimization problem that assigns a bit width and a LoRA rank to each layer for fine-tuning. Solving this problem is difficult: the search space is large and fully discrete, making exhaustive enumeration infeasible. More importantly, low-cost proxies are unreliable because they do not model the interaction between quantization noise and adapter updates \citep{frantar2023gptq,zhao2025benchmarkingPTQ}. Reliable evaluation therefore requires at least partial fine-tuning, which turns the search into an expensive black-box optimization problem. In many deployment settings, this search is run offline and its cost can be amortized over repeated deployments, making search time a secondary constraint relative to memory and post-fine-tuning performance. Nevertheless, repeated trial-and-error via fine-tuning remains prohibitively expensive.

To this end, we adopt a multi-fidelity search strategy that quickly filters poor configurations using short fine-tuning runs, and allocates longer fine-tuning runs only to a small set of promising candidates. We propose \textbf{AutoQRA}, a coarse-to-fine framework for automated quantization and rank allocation.
AutoQRA uses a two-phase design that balances global coverage with local refinement.

In \emph{Phase I}, AutoQRA performs a global multi-fidelity evolutionary search to approximate the Pareto frontier over accuracy and memory.
The population is warm-started with layer-wise importance priors, and importance-guided mutations focus edits on influential layers.
A learned surrogate model screens candidates and improves promotion decisions \citep{ru2020cocabo}, while the returned frontier is formed from real measured evaluations rather than surrogate predictions. 
In \emph{Phase II}, AutoQRA refines strong Phase I candidates with trust-region Bayesian optimization \citep{turbo}.
We fit a Gaussian process surrogate on evaluations at the highest-fidelity setting and select configurations using Expected Improvement (EI) \citep{jones1998ego}.
Both phases terminate automatically when improvements saturate, using hypervolume progress for Phase I \citep{zitzler1999hypervolume} and acquisition saturation for Phase II.

Our contributions are: (1) We formulate joint per layer bit width and LoRA rank allocation under a strict memory budget, and explain why decoupled pipelines are misaligned with post fine tuning performance. (2) We introduce AutoQRA, a two phase coarse to fine framework that combines multi fidelity evolutionary screening with trust region Bayesian refinement to search the discrete joint space efficiently. (3) Experiments show that AutoQRA achieves performance close to full precision fine tuning with a memory footprint comparable to uniform 4 bit methods.
\section{Related Work}
\label{sec:relate}

Our work is situated at the intersection of parameter-efficient fine-tuning and automated model compression, building upon advances in quantization and black-box optimization.

\subsection{Efficient LLM Fine-Tuning}

\textbf{Quantization and Mixed-Precision.}
PTQ serves as a foundation for compressing LLMs, with methods like GPTQ \citep{frantar2023gptq} and AWQ \citep{lin2023awq} utilizing second-order information or activation statistics to minimize reconstruction error. While highly effective for inference, standard PTQ often degrades performance when weights are frozen during subsequent fine-tuning. To address layer-wise sensitivity, mixed-precision techniques such as SliM-LLM \citep{huang2025slimllm} allocate bit-widths based on Hessian spectra or salience metrics. However, these methods typically focus exclusively on weight precision, treating the adaptation capacity as a fixed constant or ignoring it entirely.

\textbf{PEFT.}
Exemplified by LoRA \citep{hu2022lora}, freezes the backbone and injects trainable low-rank matrices. QLoRA \citep{dettmers2023qlora} further democratized access by quantizing the backbone to 4-bit, yet it retains a uniform rank assignment. Acknowledging that uniform capacity is inefficient, adaptive approaches like AdaLoRA \citep{zhang2023adalora} and RankAdaptor \citep{zhou2025rankadaptor} dynamically prune or allocate ranks based on singular value importance. Crucially, these methods optimize the \emph{topology} of adapters but assume a static, uniform precision for the underlying weights, failing to exploit the memory trade-offs available through variable quantization. Recent efforts have attempted to bridge these two paradigms. LoftQ \citep{li2023loftq} and LQ-LoRA \citep{guo2023lq} propose alternating optimization schemes or initialization heuristics to align quantization with low-rank structures. While pioneering, they rely on localized proxies (e.g., reconstruction loss) rather than global task performance, and they typically resort to iterative heuristics rather than a principled global search over the joint discrete design space.

\begin{figure*}[t]
    \centering
    \begin{subfigure}[b]{0.49\linewidth}
        \centering
        \includegraphics[width=\linewidth]{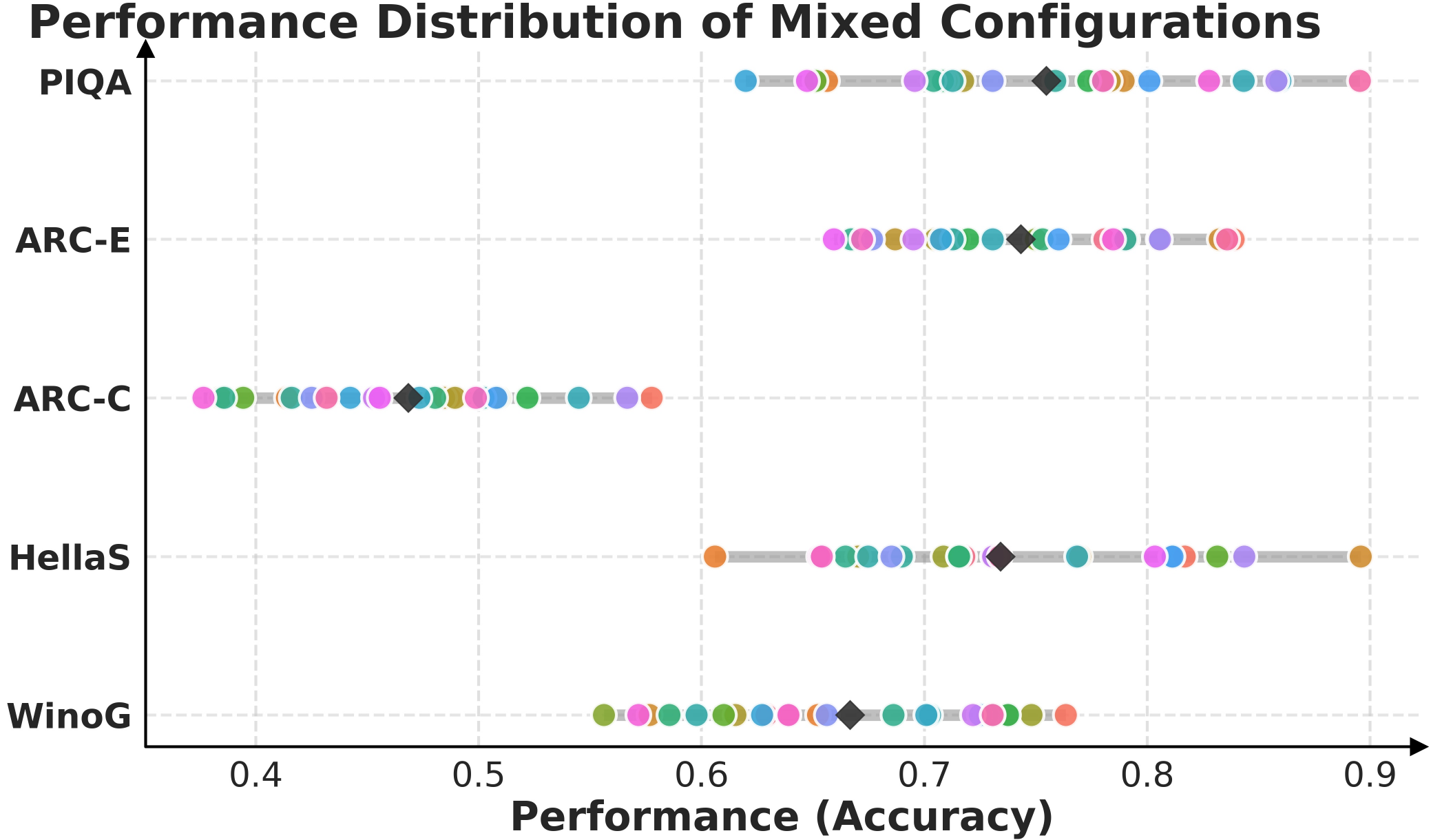}
        \caption{\textbf{Sensitivity Analysis}}
        \label{fig:variance}
    \end{subfigure}
    \hfill
    \begin{subfigure}[b]{0.49\linewidth}
        \centering
        \includegraphics[width=\linewidth]{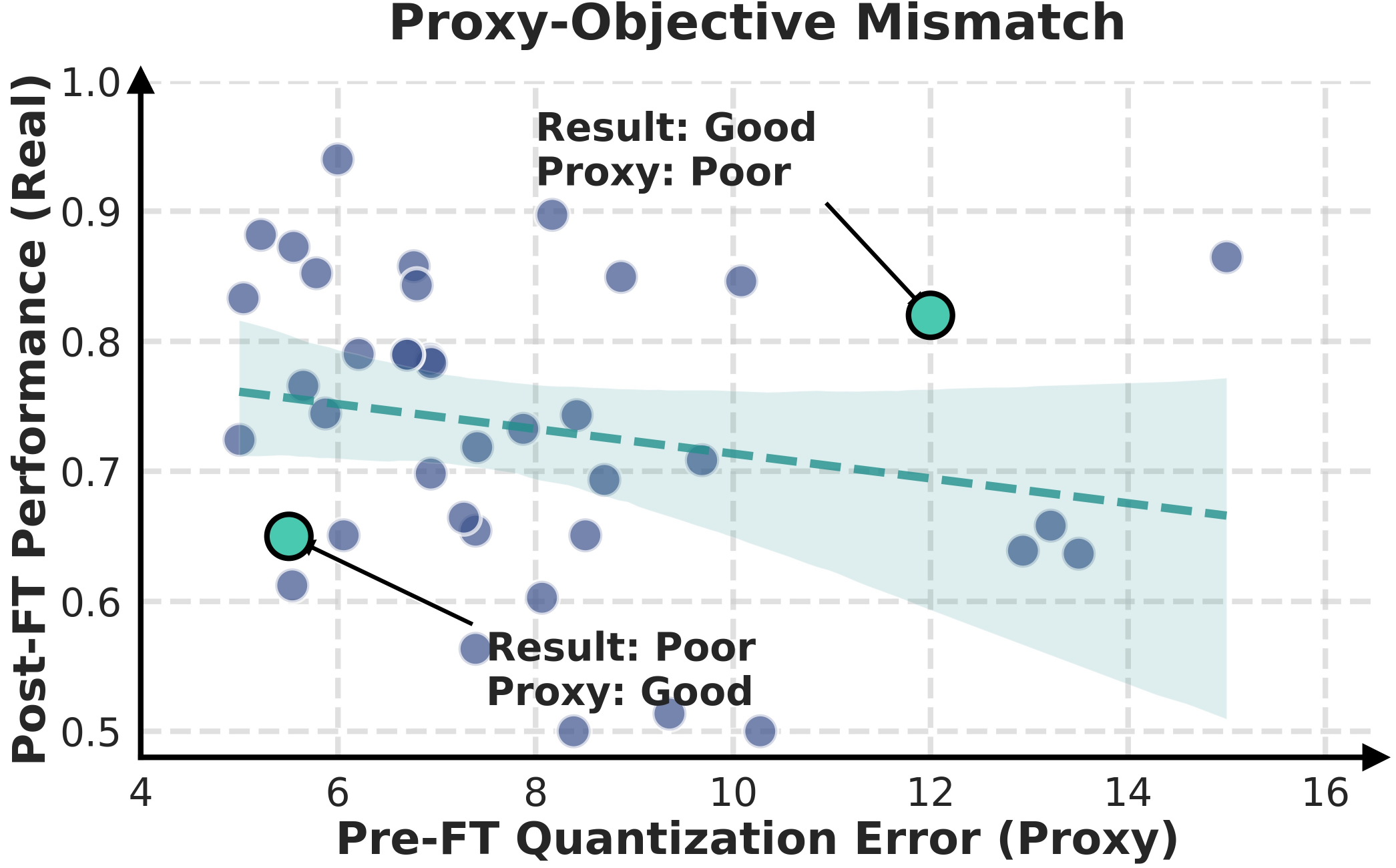}
        \caption{\textbf{Proxy Failure}}
        \label{fig:proxy_mismatch}
    \end{subfigure}
    
    \vspace{-5pt}
    \caption{\textbf{Empirical Motivation for Joint Optimization.} 
    \textbf{(a) Impact of Joint Allocation:} We visualize the accuracy distribution of feasible mixed-precision configurations across tasks. The substantial performance spread demonstrates that distinct pairings of bit-width ($q$) and rank ($r$) yield vastly different outcomes even under the same memory budget.
    \textbf{(b) Proxy-Objective Mismatch:} Standard calibration metrics (Perplexity, x-axis) fail to predict post-finetuning accuracy (y-axis). The weak correlation ($\rho{=}0.46$) and frequent rank reversals indicate that static proxies cannot reliably identify configurations where learnable adapters compensate for quantization noise.}
    \label{fig:motivation_combined}
    \vspace{-15pt}
\end{figure*}

\subsection{Automated Search for Model Compression}

Our approach also draws inspiration from Neural Architecture Search and Automated Machine Learning, which frame compression as a discrete optimization problem.

\textbf{Search Strategies for Compression.}
Early works in NAS utilized Reinforcement Learning (RL) or Evolutionary Algorithms (EAs) to discover efficient architectures \citep{zoph2016neural, real2019regularized}. For model compression, similar search strategies have been applied to find per-layer quantization policies \citep{wang2019haq}. However, these methods often incur prohibitive computational costs, making them impractical for LLM fine-tuning loops.

\textbf{Sample-Efficient Black-Box Optimization.}
To mitigate search costs, multi-fidelity optimization has emerged as a standard. Hyperband \citep{li2018hyperband} and BOHB \citep{falkner2018bohb} leverage cheap, low-fidelity approximations (e.g., partial epochs) to efficiently allocate resources to promising candidates. Techniques for optimizing over mixed categorical and continuous spaces, such as CoCaBO \citep{ru2020cocabo}, demonstrate that leveraging correlations between variables can significantly accelerate convergence. 
\newcommand{\CNSGA}{\textsc{ConstrainedNSGA2}}

\begin{figure*}[ht]
  \centering
  \includegraphics[width=0.9\linewidth]{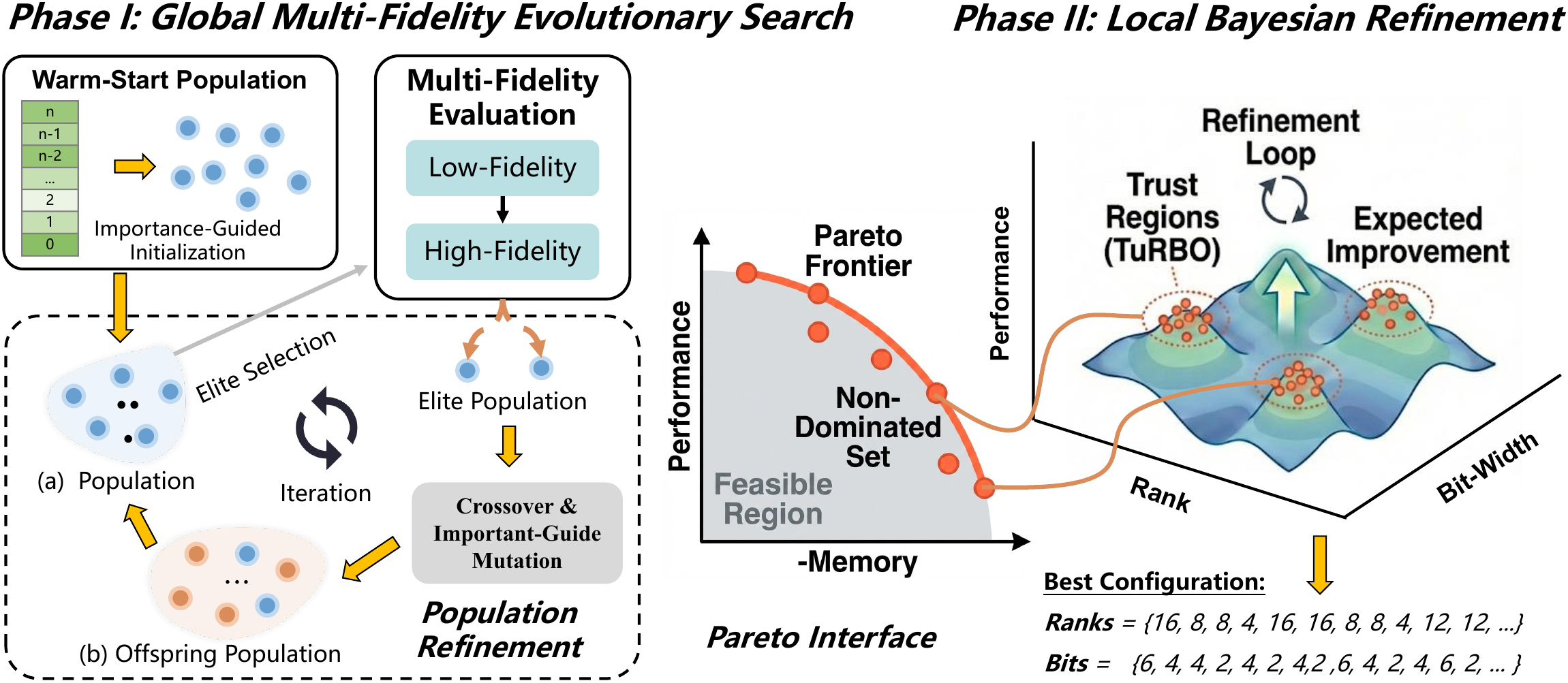}
  \caption{\textbf{Overview of the AutoQRA framework.} \textbf{Phase I} (left) approximates the global Pareto frontier via a multi-fidelity evolutionary search, utilizing importance-guided mutations and surrogate screening to navigate the discrete space. \textbf{Phase II} (right) performs a local Bayesian refinement to identify a precise operating point that maximizes user utility under the budget constraint.}
  \label{fig:framework_overview}
  \vspace{-12pt}
\end{figure*}

\section{Methodology}
\label{sec:methodology}

We introduce \textbf{AutoQRA} (\underline{Auto}mated \underline{Q}uantization--\underline{R}ank \underline{A}llocation), a framework designed to resolve the evaluation dilemma in joint model compression and adaptation. AutoQRA formulates the allocation of bit-width and rank as a constrained black-box optimization problem, replacing unreliable static proxies with dynamic assessment. To navigate the prohibitive search space efficiently, we adopt a \emph{coarse-to-fine} strategy: Phase I approximates the global Pareto frontier via a multi-fidelity evolutionary search, while Phase II performs local Bayesian refinement to pinpoint configurations where adapter capacity maximally compensates for quantization noise (Figure~\ref{fig:framework_overview}).

\subsection{Motivation and Problem Formulation}
\label{subsec:problem}

\textbf{Motivation}
We identify a critical dependency between quantization precision and adaptation rank. Figure~\ref{fig:variance} illustrates the performance distributions of diverse mixed bit-rank configurations across five downstream tasks. Crucially, distinct combinations of bit-width ($q$) and rank ($r$) yield vastly different outcomes even under identical memory constraints. We observe substantial performance fluctuations, with accuracy gaps exceeding 25\% on tasks such as Winogrande and ARC-Challenge. The consistent relative performance of specific configurations, indicated by color markers, demonstrates that downstream utility is strictly coupled with the joint allocation of $q$ and $r$. Suboptimal combinations lead to severe degradation, whereas optimal pairings approach full-precision performance.
This volatility underscores that bit-width and rank cannot be optimized in isolation. Static metrics employed by decoupled methods fail to capture these non-linear interactions. We also provide a detailed discussion on the orthogonality between backbone and adapter contributions in Appendix~\ref{app:Orthogonal}.

A common strategy is to guide bit allocation using PTQ-style calibration metrics.
However, once adapters are introduced, static proxies become unreliable for selecting learnable joint allocations.
We randomly sample $n{=}30$ feasible bit--rank configurations and compare a calibration proxy
(neg.\ log perplexity computed with frozen weights) to the final post-finetuning task score.
Figure~\ref{fig:proxy_mismatch} shows only moderate correlation and frequent rank reversals.
Two failure modes are salient.
First, configurations with worse proxy values can still fine-tune well when rank is assigned to learnable layers that compensate quantization noise.
Second, even when proxy values are similar, post-finetune accuracy can vary substantially across configurations,
indicating that proxy-matched quantization error does not determine fine-tuning outcomes.
Together with Figure~\ref{fig:variance}, this mismatch motivates treating joint bit--rank allocation as a constrained black-box optimization problem.

\textbf{Discussion.}
A key reason for the mismatch between quantization proxies and downstream fine tuning performance is that common proxies evaluate forward reconstruction of a frozen backbone, whereas downstream adaptation is an optimization problem carried out in a constrained update space. As a result, a bit width allocation that appears favorable under reconstruction or calibration criteria can still lead to poor fine tuned performance, because it may introduce error patterns that are misaligned with the correction directions expressible by low rank updates. Equivalently, the per layer bit pattern imposes a structural constraint on the function class of the frozen backbone, and only a subset of these constrained structures remains highly trainable under a fixed rank budget. This viewpoint is consistent with findings related to the lottery ticket hypothesis \citep{frankle2019lottery, sta_lottery}, which suggest that under strong structural constraints only certain parameterizations or subnetworks remain highly trainable and can reach high accuracy after optimization. In our setting, different bit allocations correspond to different constrained structures, while rank allocation controls the degrees of freedom available to compensate quantization noise during training. 

\textbf{Design Space and Budget.}
Consider a transformer model with $L$ layers. We define discrete search spaces for bit-widths $\mathcal{Q}$ (e.g., $\{2, 4, 8\}$) and LoRA ranks $\mathcal{R}$ (e.g., $\{4, 8, 16\}$). A configuration $C$ is a sequence of tuples specifying the precision and topology for each layer:
\begin{equation}
C = \{(q_\ell, r_\ell)\}_{\ell=1}^L, \quad \text{where } q_\ell \in \mathcal{Q}, \; r_\ell \in \mathcal{R}.
\end{equation}
The total memory footprint $M(C)$ comprises the quantized backbone, trainable adapters, and quantization metadata. Let $N(W_\ell)$ be the parameter count of the dense weights at layer $\ell$. Assuming adapters utilize 16-bit precision ($p_r=16$), the layer-wise memory cost is:
\begingroup\small
\begin{equation}
\label{eq:mem}
m_\ell(C) = \underbrace{\frac{N(W_\ell) \cdot q_\ell}{8} + m^{\mathrm{meta}}_\ell}_{\text{Quantized Backbone}} + \underbrace{\frac{(d_{\text{in}} r_\ell + r_\ell d_{\text{out}}) \cdot p_r}{8}}_{\text{Trainable Adapters}},
\end{equation}
\endgroup
where $m^{\mathrm{meta}}_\ell$ accounts for quantization scales and zero-points. We aim to maximize the validation performance $P(C)$ after fine-tuning, subject to a strict global memory budget $B_{\text{max}}$.

\textbf{The Optimization Challenge.}
Formally, we solve the constrained optimization problem:
\begin{equation}
\label{eq:opt_problem}
\begin{aligned}
& \underset{C}{\text{maximize}}
& & P(C) \coloneqq \text{Metric}(\text{FineTune}(C)) \\
& \text{subject to}
& & \sum_{\ell=1}^L m_\ell(C) \le B_{\text{max}}.
\end{aligned}
\end{equation}
This formulation presents two challenges: (1) $P(C)$ is a \emph{black-box} function with no closed-form gradient relative to discrete $q_\ell$ or $r_\ell$; and (2) evaluating $P(C)$ is computationally expensive, prohibiting exhaustive search over the exponential space $(|\mathcal{Q}| \times |\mathcal{R}|)^L$.

Phase I approximates the Pareto frontier $\mathcal{C}^*$ of non-dominated configurations. Phase II (Refinement) subsequently maximizes a scalarized utility function $f(C; \alpha)$ governed by a user preference parameter $\alpha \in [0,1]$:
\begin{equation}
\label{eq:scalarize}
f(C; \alpha) = \alpha \cdot \hat{P}(C) - (1-\alpha) \cdot \hat{M}(C),
\end{equation}
where $\hat{P}$ and $\hat{M}$ denote min-max normalized metrics.

\subsection{Phase I: Global Multi-Fidelity Evolutionary Search}
\label{subsec:global}

\newcommand{\Repair}{\textsc{Repair}}

Phase~I aims to construct a diverse set of \emph{feasible} bit--rank configurations that concentrate near the performance--memory Pareto frontier,
which Phase~II subsequently refines.
As illustrated in Figure~\ref{fig:framework_overview}, we combine (i) an evolutionary loop \citep{deb2002fast} for exploration and diversity with
(ii) a Hyperband-style multi-fidelity evaluation schedule, so that evaluations at the highest-fidelity setting (largest step count) are reserved for only a small fraction of candidates.

We first define the footprint used to enforce feasibility.
Following Eq.~\eqref{eq:mem}, the total footprint is
\begin{equation}
\label{eq:phase1_mem_total}
M(C)=\sum_{\ell=1}^L m_\ell(C).
\end{equation}
In all experiments, the feasibility check $M(C)\le B_{\max}$ uses the \emph{exact} accounting from our implementation, including quantization metadata
(e.g., scales and zero-points).

We use two signals to guide \emph{joint} search: $I_q(\ell)$ for proposing bit edits and $I_r(\ell)$ for proposing rank edits.
They are separated because quantization sensitivity and adaptation learnability are often mismatched, not because the optimization is decoupled.
$I_q(\ell)$ measures a layer's sensitivity to low-bit perturbations on a small calibration set, while $I_r(\ell)$ measures how much update energy a layer exhibits
during fine-tuning.
Both signals are used only for warm start and for defining proposal distributions in our operators; they are not used as substitutes for the final fine-tuning metric.See Appendix~\ref{app:phase1_details} for $I_q(\ell)$ and $I_r(\ell)$ definitions.

We encode each configuration by an ordinal embedding $\psi(C)\in\mathbb{R}^{2L}$:
each $q_\ell\in\mathcal{Q}$ and $r_\ell\in\mathcal{R}$ is mapped to its ordinal index on the corresponding ladder and then standardized,
and we concatenate all layers.

\textbf{Warm start.}
Initialization starts from an importance-shaped prototype and adds small perturbations for diversity:
\begin{equation}
\label{eq:warmstart_qr}
q_\ell^{(0)}=\big\lfloor \tau_q(\widetilde{I_q}(\ell)) \big\rceil_{\mathcal{Q}},
\qquad
r_\ell^{(0)}=\big\lfloor \tau_r(\widetilde{I_r}(\ell)) \big\rceil_{\mathcal{R}}.
\end{equation}
Here $\widetilde{I_q},\widetilde{I_r}$ denote normalized scores; $\tau_q,\tau_r$ are monotone mappings from scores to discrete ladders
that respect the hard constraint $M(C)\le B_{\max}$ (implemented via a greedy fill under the exact memory accounting).
The initial population $\mathcal{P}_0$ is formed by this prototype plus perturbed variants, followed by feasibility repair.

\textbf{Feasibility repair.}
We use a deterministic projection $\Repair(\cdot)$ to map any infeasible candidate back into the feasible set; the operator is reused in Phase~II.
Let $q_\ell^{-}$ (resp.\ $r_\ell^{-}$) denote the nearest lower choice in $\mathcal{Q}$ (resp.\ $\mathcal{R}$).
We define the memory saved by a single discrete downgrade as
\begin{equation}
\label{eq:delta_mem}
\Delta M_t(\ell)=
\begin{cases}
M(C)-M(C\!\downarrow\! q_\ell), & t=q,\\
M(C)-M(C\!\downarrow\! r_\ell), & t=r,
\end{cases}
\end{equation}
where $t\in\{q,r\}$, and $C\!\downarrow\! q_\ell$ replaces $q_\ell$ by $q_\ell^{-}$ (analogously for $r_\ell$).
While $M(C)>B_{\max}$, we apply the downgrade that minimizes sensitivity per saved memory:
\begingroup\small
\begin{equation}
\label{eq:repair_rule}
(\ell^\star, t^\star)=
\arg\min_{\ell,\; t\in\{q,r\}}
\frac{\widetilde{I_t}(\ell)+\epsilon}{\Delta M_t(\ell)}
\quad \text{s.t.}\quad \Delta M_t(\ell)>0,
\end{equation}
\endgroup
with $I_q$ used for $t{=}q$ and $I_r$ used for $t{=}r$.
This rule removes capacity from low-sensitivity layers and from the knob (bit or rank) that yields the largest memory relief
for the smallest expected damage, terminating once feasibility is restored.

\textbf{Variation operators.}
Offspring are generated by two complementary mutations, each followed by $\Repair(\cdot)$.
A sensitivity-guided mutation selects a layer according to
\begin{equation}
\begin{split}
\Pr_t(\ell) &= \frac{I_t(\ell)^\gamma}{\sum_{j=1}^L I_t(j)^\gamma},
\qquad t\in\{q,r\},\\
&\hspace{0.5em}\text{(}t=q\text{: bit updates;\ } t=r\text{: rank updates)\,}.
\end{split}
\end{equation}
where $\gamma>0$ controls the concentration of importance sampling.
It then changes $q_\ell$ or $r_\ell$ by one adjacent discrete step.
A memory-balanced coupled mutation applies a primary memory-increasing edit (e.g., $q_{\ell}\!\uparrow$ or $r_{\ell}\!\uparrow$),
then performs compensating memory-decreasing edits (possibly on different layers) until feasibility is restored.
This global compensation avoids the scale mismatch of within-layer ``iso-memory'' pairing and keeps the search concentrated near the constraint boundary.

\textbf{Multi-fidelity evaluation.}
We parameterize evaluation cost by the number of fine-tuning steps $T$ (equivalently, a fractional number of epochs) and use an increasing ladder
$0<T_1<\cdots<T_S$ with pruning factor $\eta>1$.
Evaluations at smaller $T_s$ provide inexpensive but noisier estimates $P(C;T_s)$ (low-fidelity, LF), while $T_S$ denotes the \emph{largest step count used in our search} (high-fidelity, HF) and is not assumed to correspond to fully converged fine-tuning.
When a candidate is promoted from $T_s$ to $T_{s+1}$, it \emph{continues training from its checkpoint at $T_s$} rather than restarting,
so that $P(C;T_{s+1})$ is directly comparable across candidates.
To reduce variance in LF ranking, we fix the optimizer hyperparameters, random seed, and data order across candidates.

\textbf{Surrogate screening.}
Multi-fidelity screening is used exclusively for promotion decisions.
At stage $s$, we train $\Phi_s$ to predict the HF score $P(C;T_S)$ from LF observations.
Let $\mathcal{D}_s=\{(C_i, P(C_i;T_s), P(C_i;T_S))\}$ be the paired set collected so far (across generations).
With feature vector $x_i=[P(C_i;T_s), \log M(C_i), \psi(C_i)]$, we fit $\Phi_s(\cdot;\theta_s)$ by regression:
\begin{equation}
\label{eq:screen_loss}
\theta_s^\star=
\arg\min_{\theta}
\sum_{(C_i,\cdot)\in \mathcal{D}_s}
\rho\!\left(\Phi_s(x_i;\theta)-P(C_i;T_S)\right)
+\lambda\|\theta\|_2^2,
\end{equation}
where $\rho(\cdot)$ is the Huber loss to mitigate occasional LF outliers.
When $|\mathcal{D}_s|$ is insufficient early on, promotion falls back to ranking by the measured $P(C;T_s)$.
Crucially, Pareto selection uses \emph{measured} $P(C;T_S)$ rather than surrogate predictions, ensuring the returned frontier is supported by
evaluations at $T_S$.

\textbf{Update and termination.}
After each generation, we update the population using NSGA-II with constrained domination:
feasible candidates dominate infeasible ones; among infeasible candidates, lower constraint violation is preferred.
The loop terminates when the dominated hypervolume(Appendix~\ref{app:Hypervolume}) of the feasible front stabilizes
(relative improvement below $\epsilon_{\mathrm{hv}}$ for $\Delta$ generations), yielding a compact feasible Pareto set for Phase~II.

\subsection{Phase II: Local Bayesian Refinement}
\label{subsec:local}

Phase~I is primarily responsible for \emph{coverage}: it explores the combinatorial bit--rank space under the hard constraint
$M(C)\le B_{\max}$ and returns a diverse set of \emph{measured} candidates evaluated at $T_S$ near the Pareto frontier.
Phase~II is responsible for \emph{selection and sharpening}: given a user preference $\alpha$ (Eq.~\eqref{eq:scalarize}),
we identify a single operating point by improving the scalarized utility $f(C;\alpha)$ using a small number of additional
evaluations at $T_S$.

A natural concern is that a handful of Phase~II evaluations may be insufficient in the original exponential space.
We therefore do \emph{not} optimize over the full space in Phase~II.
Instead, Phase~II operates on a reduced search region that Phase~I has already verified to be feasible and competitive at $T_S$.
Moreover, rather than committing to a single local basin, we follow the multi-region trust-region principle of TuRBO \citep{turbo}:
we maintain several discrete trust regions around multiple strong Phase~I solutions and allocate the limited Phase~II evaluations to the region
with the highest predicted improvement.

Let $\mathcal{D}^{\mathrm{hi}}=\{(C_i,y_i)\}_{i=1}^{n_0}$ collect all \emph{measured} evaluations at $T_S$ after Phase~I.
Each $C_i$ is evaluated at $T_S$ and satisfies $M(C_i)\le B_{\max}$.

For each entry we record
\begingroup\small
\begin{equation}
\label{eq:phase2_obs}
\begin{aligned}
y_i &\coloneqq f(C_i;\alpha) \coloneqq \alpha\,\hat{P}(C_i)-(1-\alpha)\,\hat{M}(C_i), \\
&\quad P(C_i) \coloneqq P(C_i;T_S)
\end{aligned}
\end{equation}
\endgroup
We set $y_0^+=\max_{1\le i\le n_0} y_i$ and initialize $J$ trust-region centers by taking the top candidates in $\mathcal{D}^{\mathrm{hi}}$
while enforcing diversity in atomic distance:
we greedily select $C^{(1)}_0,\dots,C^{(J)}_0$ such that each new center satisfies
$d_{\mathrm{atom}}(C^{(j)}_0,C^{(j')}_0)\ge \Delta_{\mathrm{div}}$ for all $j'<j$.
The global incumbent is the best measured one in $\mathcal{D}^{\mathrm{hi}}$ (ties broken by smaller $M(C_i)$).

We model the utility landscape with a Gaussian process surrogate in the standardized ordinal embedding $\psi(C)\in\mathbb{R}^{2L}$ used throughout.
We place a GP prior on a latent function $g(C)$ and assume noisy observations
\begingroup\small
\begin{equation}
\label{eq:phase2_gp}
g(C)\sim\mathcal{GP}\!\big(0,k(C,C')\big),\ y=g(C)+\varepsilon,\ \ \varepsilon\sim\mathcal{N}(0,\sigma_n^2),
\end{equation}
\endgroup
refitting hyperparameters by maximizing the GP marginal likelihood on the updated $\mathcal{D}^{\mathrm{hi}}$.
We use a Mat\'ern-$5/2$ kernel in the embedded space:
\begin{equation}
\label{eq:phase2_kernel}
\begin{aligned}
k(C,C')&=\sigma_f^2\!\left(1+\sqrt{5}\,r+\frac{5}{3}r^2\right)\exp(-\sqrt{5}\,r),\\
r&=\|\psi(C)-\psi(C')\|_2/\ell_{\mathrm{gp}}.
\end{aligned}
\end{equation}

Phase~II proposes candidates only within discrete trust regions.
An \emph{atomic edit} changes exactly one variable to an adjacent value on its ladder (either a neighbor in $\mathcal{Q}$ for $q_\ell$
or a neighbor in $\mathcal{R}$ for $r_\ell$).
Let $d_{\mathrm{atom}}(C,C')$ be the minimum number of atomic edits needed to transform $C$ into $C'$.
For region $j$ at iteration $t$, with center $C^{(j)}_t$ and integer radius $\delta_{j,t}$, we first define the pre-repair neighborhood
\begin{equation}
\label{eq:phase2_ball}
\mathcal{B}_{j,t}\triangleq
\big\{\, C' \ \big|\ d_{\mathrm{atom}}(C',\, C^{(j)}_t)\le \delta_{j,t} \,\big\},
\end{equation}
and then project every candidate to feasibility using the same operator $\Repair(\cdot)$ as in Phase~I:
\begin{equation}
\label{eq:phase2_tr}
\Omega_{j,t}\triangleq
\big\{\, \Repair(C') \ \big|\ C'\in \mathcal{B}_{j,t}\,\big\}.
\end{equation}
Thus every $C\in\Omega_{j,t}$ satisfies $M(C)\le B_{\max}$ under the same exact accounting as Phase~I.
We then form the union $\Omega_t=\bigcup_{j=1}^J \Omega_{j,t}$.
When $\Omega_t$ is too large to enumerate, we subsample a fixed-size pool from each neighborhood before applying $\Repair(\cdot)$, using the same sensitivity-guided proposal distribution as in Phase~I.

\begin{table*}[t]
\centering
\scriptsize
\setlength{\tabcolsep}{3.2pt}
\caption{\textbf{Main results across four backbones.}
We report task-average accuracy (\%), average weight precision (AvgBit), average LoRA rank (AvgRank), and total memory footprint (Mem, GB).
Bold / underline indicate best / second best per backbone.}
\label{tab:main_all_models}
\resizebox{\linewidth}{!}{
\begin{tabular}{lcccccccccccccccc}
\toprule
\multirow{2}{*}{\textbf{Method}}
& \multicolumn{4}{c}{\textbf{LLaMA3.1-8B}}
& \multicolumn{4}{c}{\textbf{LLaMA3.2-3B}}
& \multicolumn{4}{c}{\textbf{Qwen2.5-7B}}
& \multicolumn{4}{c}{\textbf{Qwen2.5-3B}} \\
\cmidrule(lr){2-5}\cmidrule(lr){6-9}\cmidrule(lr){10-13}\cmidrule(lr){14-17}
& Avg$\uparrow$ & AvgBit$\downarrow$ & AvgRank & Mem$\downarrow$
& Avg$\uparrow$ & AvgBit$\downarrow$ & AvgRank & Mem$\downarrow$
& Avg$\uparrow$ & AvgBit$\downarrow$ & AvgRank & Mem$\downarrow$
& Avg$\uparrow$ & AvgBit$\downarrow$ & AvgRank & Mem$\downarrow$ \\
\midrule
LoRA (FP16) & \underline{69.94} & 16.00 & 16.00 & 20.50 & 65.40 & 16.00 & 16.00 & 10.40 & 71.33 & 16.00 & 16.00 & 20.20 & 65.53 & 16.00 & 16.00 & 10.84 \\
QLoRA (4-bit) & 67.45 & 4.00 & 16.00 & 15.22 & 64.43 & 4.00 & 16.00 & 8.90 & 69.01 & 4.00 & 16.00 & 15.60 & 62.89 & 4.00 & 16.00 & 8.16 \\
AdaLoRA (4-bit) & 66.36 & 4.00 & 15.84 & 14.92 & 61.88 & 4.00 & 15.73 & 8.60 & 69.05 & 4.00 & 15.65 & 15.40 & 62.51 & 4.00 & 15.92 & 8.70 \\
LoftQ (4-bit) & 68.65 & 4.00 & 16.00 & 15.13 & 64.91 & 4.00 & 16.00 & 8.83 & 69.35 & 4.00 & 16.00 & 15.33 & 62.71 & 4.00 & 16.00 & 8.21 \\
LQ-LoRA & 67.82 & 3.78 & 16.00 & 20.12 & 63.63 & 3.85 & 16.00 & \underline{7.12} & 66.91 & 3.85 & 16.00 & 18.52 & 62.86 & \textbf{3.63} & 16.00 & \underline{7.05} \\
AMQ+LoRA & 67.75 & \underline{3.88} & 16.00 & 14.45 & 63.31 & \underline{3.91} & 16.00 & 7.22 & 70.71 & \underline{3.91} & 16.00 & 14.90 & 62.81 & 4.00 & 16.00 & 8.17 \\
AMQ+AdaLoRA & 67.63 & 3.93 & \textbf{10.18} & \underline{14.40} & 63.38 & 3.96 & \textbf{9.68} & 7.25 & 70.66 & 3.96 & \textbf{10.18} & \underline{14.70} & 64.88 & 3.80 & 15.84 & 8.65 \\
\midrule
AutoQRA ($\le$4bit) & 69.83 & \textbf{3.75} & \underline{10.50} & \textbf{13.08} & \underline{65.58} & \textbf{3.64} & \underline{11.14} & \textbf{6.72} & \underline{71.35} & \textbf{3.71} & \underline{10.57} & \textbf{11.95} & \underline{66.33} & \underline{3.72} & \textbf{9.78} & \textbf{6.45} \\
AutoQRA (optimal) & \textbf{70.45} & 5.25 & 12.25 & 17.32 & \textbf{66.16} & 5.14 & 12.57 & 8.41 & \textbf{73.19} & 5.36 & 12.70 & 17.24 & \textbf{68.05} & 5.22 & \underline{12.00} & 8.31 \\
\bottomrule
\end{tabular}}
\vspace{-8pt}
\end{table*}

Given the GP posterior mean $\mu_t(C)$ and standard deviation $\sigma_t(C)$, we use Expected Improvement (EI)
with respect to the best measured utility
$y_t^+=\max\{y_i:(C_i,y_i)\in\mathcal{D}^{\mathrm{hi}}\}$:
\begingroup\small
\begin{equation}
\label{eq:phase2_ei}
\begin{aligned}
z_t(C) &\coloneqq \frac{\mu_t(C)-y_t^+}{\sigma_t(C)},\\
\mathrm{EI}_t(C) &\coloneqq (\mu_t(C)-y_t^+)\,\Phi\!\big(z_t(C)\big)
+\sigma_t(C)\,\phi\!\big(z_t(C)\big),
\end{aligned}
\end{equation}
\endgroup
We select the next configuration by scanning the pool $\Omega_t$:
\begin{equation}
\label{eq:phase2_next}
C_{t+1}=\arg\max_{C\in\Omega_t}\ \mathrm{EI}_t(C),
\end{equation}
evaluate it at $T_S$ to obtain $y_{t+1}=f(C_{t+1};\alpha)$, and update
$\mathcal{D}^{\mathrm{hi}}\leftarrow \mathcal{D}^{\mathrm{hi}}\cup\{(C_{t+1},y_{t+1})\}$.
Surrogate predictions are used only for proposing evaluations; the final returned configuration is always supported by measured utility.

We adapt the trust-region radius only for the region that generated $C_{t+1}$.
Let $j(t)$ denote the selected region (i.e., $C_{t+1}\in\Omega_{j(t),t}$). For every region $j\in\{1,\ldots,J\}$,
\begin{equation}
\label{eq:phase2_delta}
\delta_{j,t+1}=
\begin{cases}
\min\{\kappa_{\uparrow}\delta_{j,t},\delta_{\max}\}, & j=j(t),\ y_{t+1}>y_t^+,\\
\max\{\kappa_{\downarrow}\delta_{j,t},\delta_{\min}\}, & j=j(t),\ y_{t+1}\le y_t^+,\\
\delta_{j,t}, & j\ne j(t),
\end{cases}
\end{equation}
and we update the corresponding center by
\begin{equation}
\label{eq:phase2_center}
C^{(j)}_{t+1}=
\begin{cases}
C_{t+1}, & j=j(t),\ y_{t+1}>y_t^+,\\
C^{(j)}_{t}, & \text{otherwise}.
\end{cases}
\end{equation}

The loop terminates when either (i) no meaningful improvement is predicted,
\begin{equation}
\label{eq:phase2_stop}
\max_{C\in\Omega_t}\ \mathrm{EI}_t(C)<\epsilon_{\mathrm{ei}},
\end{equation}
or (ii) a hard cap of $N_{\max}$ Phase~II iterations is reached.
We return the best measured feasible configuration
$C^\star=\arg\max_{(C_i,y_i)\in\mathcal{D}^{\mathrm{hi}}}y_i$.
\section{Experiments}
\label{sec:experiments}

\subsection{Experimental Setup}\label{subsec:experimentalsetup}

\textbf{Datasets and LLMs.}
We fine-tune on Alpaca52k and HC3 \citep{alpaca}, and evaluate zero-/few-shot on BoolQ \citep{clark2019boolq}, PIQA \citep{bisk2020piqa}, HellaSwag \citep{zellers2019hellaswag}, WinoGrande \citep{sakaguchi2021winogrande}, ARC-E/ARC-C \citep{clark2018think}, OpenBookQA \citep{mihaylov2018can}, and MMLU \citep{hendrycks2021measuring}. Backbones include LLaMA-3.1/3.2 \citep{grattafiori2024llama3herdmodels} and Qwen-2.5 \citep{qwen2025qwen25technicalreport}.

\textbf{Baselines.}
We compare with LoRA (FP16) \citep{hu2022lora}, QLoRA (4-bit) \citep{dettmers2023qlora}, AdaLoRA \citep{zhang2023adalora}, LoftQ \citep{li2023loftq}, and LQ-LoRA \citep{guo2024lqlora}. To isolate the value of \emph{joint} bit--rank allocation, we further include a decoupled baseline that first runs an automatic
mixed-precision quantization allocator AMQ \citep{amq} and then fine-tunes with either LoRA or AdaLoRA(\textbf{AMQ+L}, \textbf{AMQ+AL}). 

\textbf{Implementation Details.}
All methods are implemented in PyTorch using the Transformers/PEFT/BitsAndBytes stack and run on NVIDIA A100 GPUs.
Methods only differ in how they allocate per-layer $(q_\ell, r_\ell)$ under the same global memory budget $B_{\max}$.
We describe the full AutoQRA search schedule (Phase~I/II step ladder, low- vs.\ high-fidelity evaluations, and early stopping) in Appendix~\ref{app:details}.

\subsection{Main Results}\label{subsec:mainresults}
Table~\ref{tab:main_all_models} reports the main results on four backbones. We summarize each method by (i) task-average accuracy and (ii) the resource triple (AvgBit, AvgRank, Mem) computed under the same memory accounting used by our implementation. Overall, AutoQRA achieves the strongest accuracy--memory trade-off: under the AvgBit$\le4$ regime, AutoQRA consistently improves over uniform 4-bit baselines (QLoRA/AdaLoRA/LoftQ) while using lower effective precision and a smaller footprint; when allowing mixed precision under the same protocol, AutoQRA (optimal) surpasses FP16 LoRA on all four backbones with substantially reduced weight precision. In particular, AutoQRA (AvgBit$\le4$) is the best-performing $\le$4-bit method across all backbones, while reducing footprint by 12--22\% relative to uniform 4-bit baselines. It also achieves these gains with markedly smaller AvgRank, highlighting the benefit of coordinated bit--rank allocation under fixed memory. Importantly, decoupled pipelines remain suboptimal under the same memory budget $B_{\max}$. AMQ+LoRA/AdaLoRA first allocates bits by a static quantization objective and only then tunes ranks, which cannot exploit the compensatory interplay between quantization noise and adapter capacity. Per-task accuracies are reported in Appendix~\ref{app:taskwise_discussion}.

\begin{figure}[t]
  \centering
  \includegraphics[width=\linewidth]{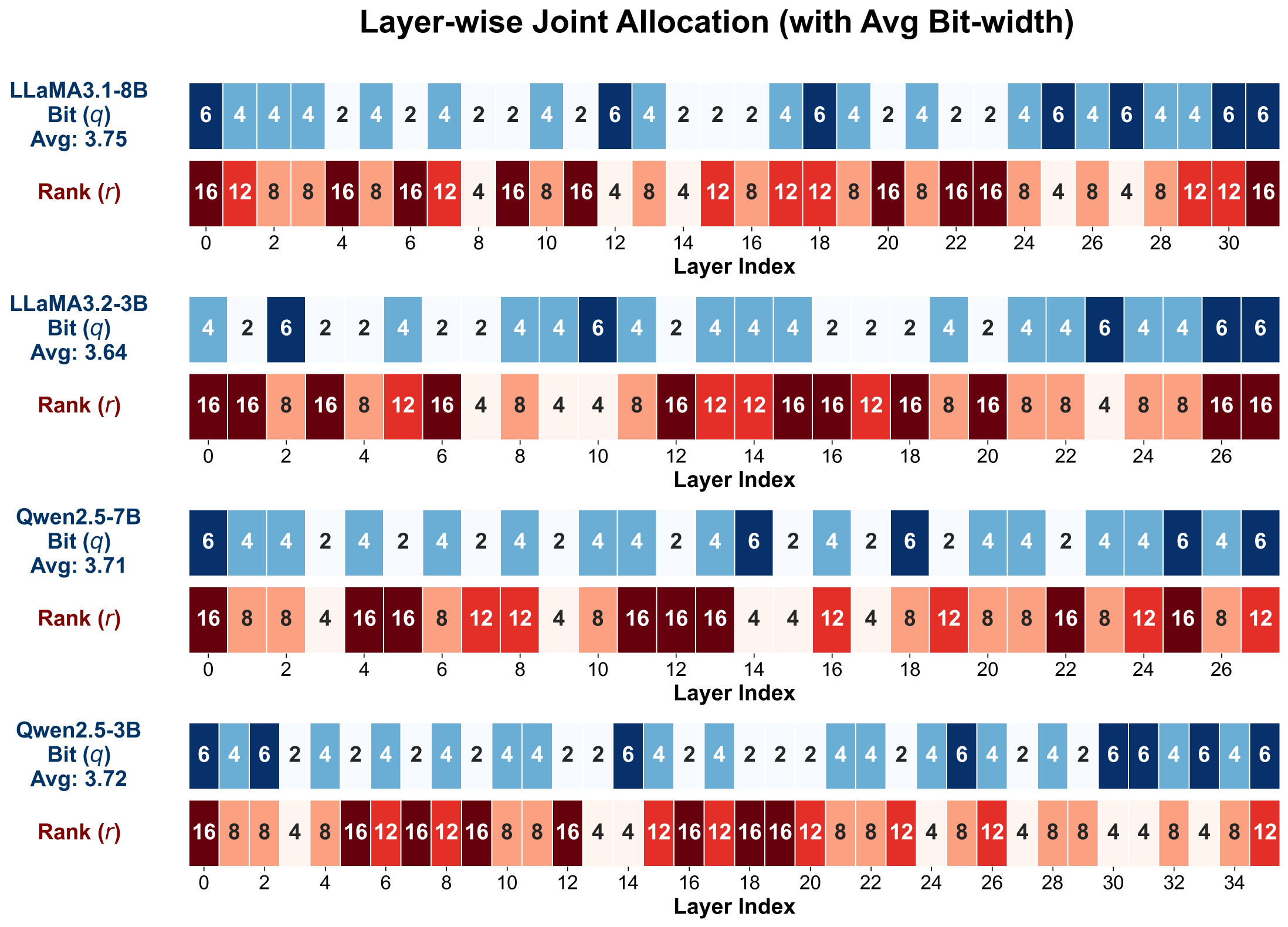}
  \vspace{-15pt}
  \caption{\textbf{Layer wise configurations found by AutoQRA show a compensation pattern.}
  Layers assigned lower bit widths are often paired with higher ranks, suggesting that adapter capacity is reallocated to compensate for quantization noise.}
  \label{fig:config_compensation}
  \vspace{-20pt}
\end{figure}

\textbf{Evidence of bit width and rank compensation.}Figure~\ref{fig:config_compensation} provides direct evidence for the compensation effect that motivates our joint search.
Across models, AutoQRA assigns high ranks to many of the layers that are quantized more aggressively, while keeping ranks small in layers that retain higher precision.
This produces a consistent negative association between $q_\ell$ and $r_\ell$ across layers, despite similar average bit widths across the compared models.
These patterns indicate that better fine tuning performance under a fixed memory budget is achieved through coordinated allocation of bit widths and ranks, rather than by optimizing quantization error alone.

\subsection{Screening surrogate quality.}
\label{subsec:surrogate}

\begin{figure}[t]
  \centering
  \includegraphics[width=\linewidth]{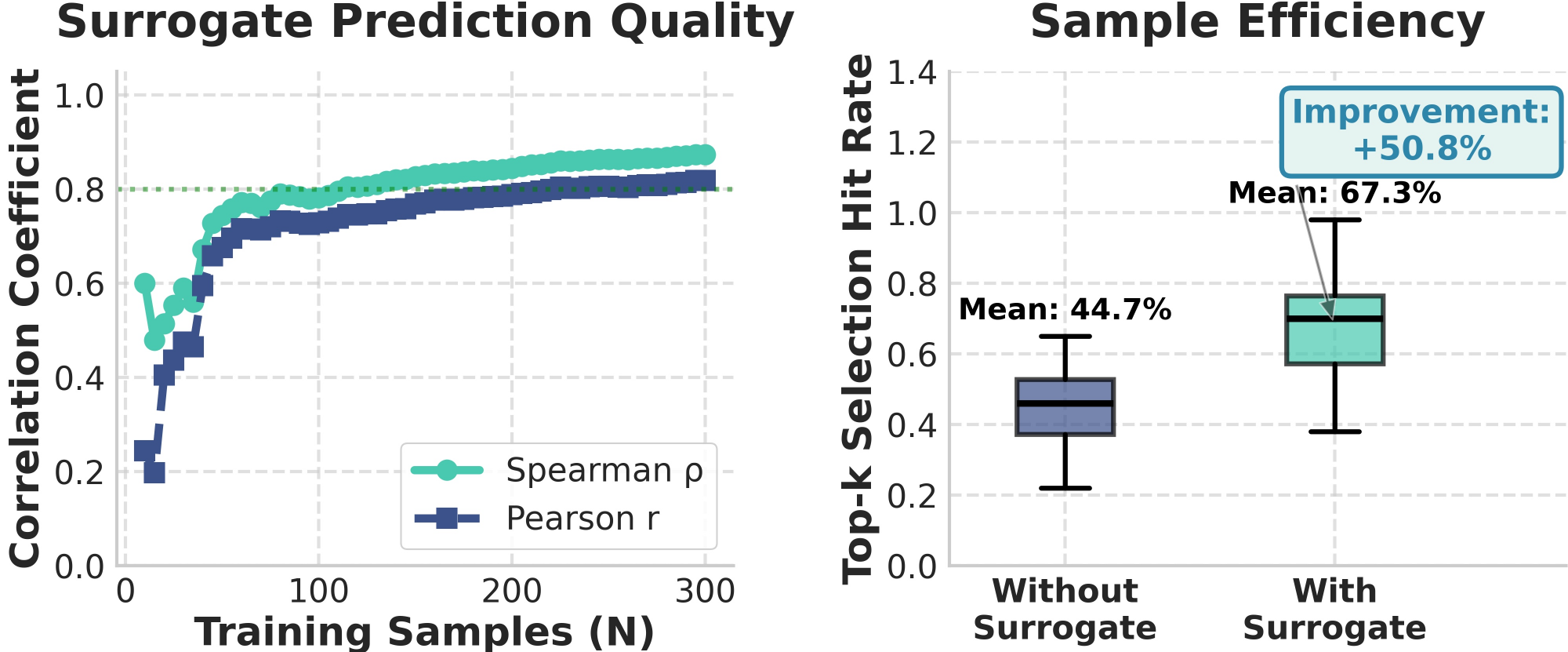}
  \vspace{-15pt}
  \caption{\textbf{Surrogate Quality.} Surrogate accuracy improves with paired data and boosts top-3 promotion hit rate.}
  \label{fig:surrogate_quality}
  \vspace{-16pt}
\end{figure}

We evaluate whether low-fidelity (LF) observations can be reliably mapped to performance at the highest-fidelity setting $T_S$ by our screening surrogate.
Figure~\ref{fig:surrogate_quality} (left) reports predictive quality as the amount of paired LF and $T_S$ data increases:
both Spearman and Pearson correlations rise quickly and then stabilize.
Figure~\ref{fig:surrogate_quality} (right) summarizes how surrogate screening affects promotion decisions.
Each boxplot aggregates the \emph{top-3 selection hit rate} over repeated trials, where the hit rate measures how often the three candidates selected for promotion align with the oracle top performers judged at $T_S$.
Surrogate-guided promotion shifts the entire distribution upward and increases the mean hit rate from 44.7\% (ranking by LF scores alone) to 67.3\%, corresponding to a 50.8\% relative improvement.

\subsection{Search efficiency and sample complexity.}
\label{subsec:search_effi}

\begin{figure}[t]
    \centering
    \includegraphics[width=\linewidth]{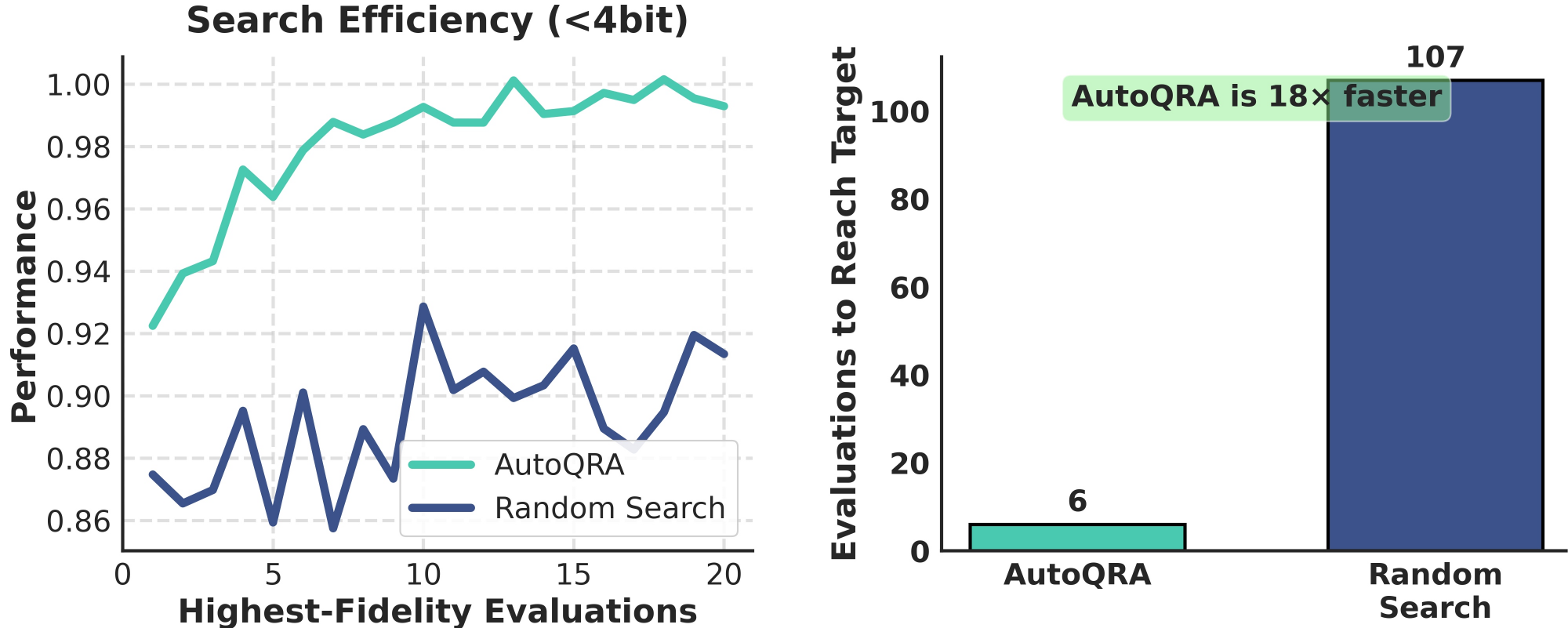}
    \vspace{-12pt}
    \caption{\textbf{Search efficiency analysis.}
    (Left) Best validation performance versus the number of evaluations at the largest search budget $b_S$.
    AutoQRA improves rapidly and consistently outperforms random search.
    (Right) Number of largest-budget evaluations required to reach a target accuracy.
    AutoQRA needs 6 evaluations compared to 107 for random search, yielding an $18\times$ reduction in expensive evaluations.}
    \label{fig:efficiency}
    \vspace{-10pt}
\end{figure}

To validate the effectiveness of our optimization strategy, we benchmark AutoQRA against random search under the same $<4$-bit memory constraint.
Figure~\ref{fig:efficiency} (left) reports the best validation performance found as a function of the number of HF evaluations at $T_S$.
AutoQRA improves rapidly, identifying strong configurations within the first few trials and then maintaining a stable trajectory.
In contrast, random search shows large variance and struggles to locate high-performing regions in the combinatorial space.
We further quantify this advantage in Figure~\ref{fig:efficiency} (right) by measuring the number of HF evaluations at $T_S$ required to reach a fixed performance target.
Random search requires 107 evaluations to reach the target, whereas AutoQRA reaches the same target with 6 evaluations. 

\subsection{Ablation Study}
\label{subsec:ablation}

We ablate key components of AutoQRA under a fixed memory budget: (i) removing the warm-start / importance prior,
(ii) disabling Phase~I global search (BO only), (iii) disabling Phase~II local refinement (EA only),
(iv) optimizing only one axis (bit-only or rank-only), and (v) removing multi-fidelity and/or surrogate screening.
Across tasks, the full system consistently yields the best accuracy--memory trade-off, while removing global exploration
or feasibility-aware search degrades the frontier most severely.

We find that without the feasibility projection $\Repair(\cdot)$, a large fraction of generated candidates violate the hard memory budget,
reducing effective search throughput and weakening Pareto coverage. Detailed diagnostics for repair behavior and
a sensitivity study of key search hyperparameters are provided in Appendix~\ref{app:additional_ablation}. 

\begin{figure}[t]
  \centering
  \includegraphics[width=0.8\linewidth]{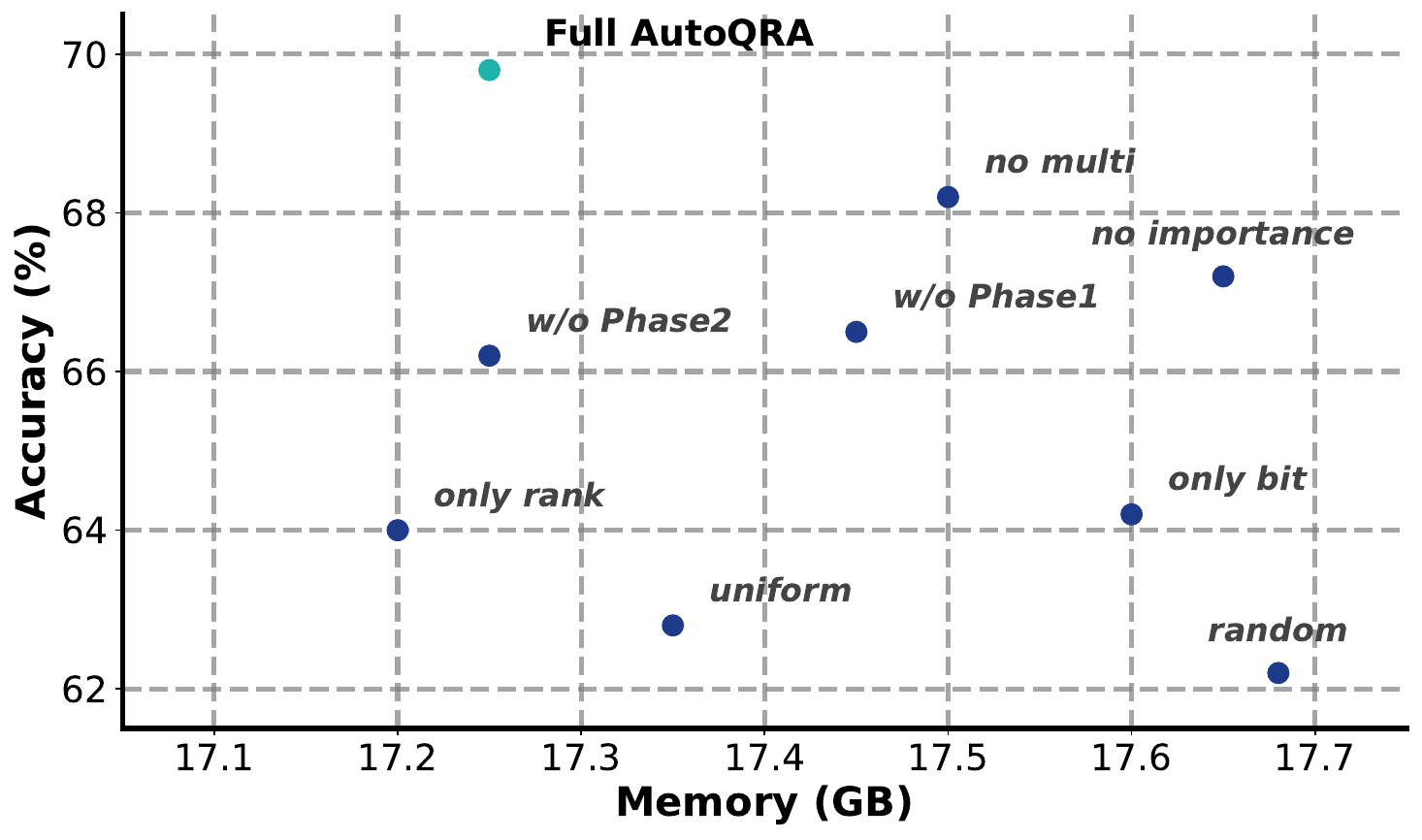}
  \vspace{-8pt}
  \caption{\textbf{Ablation.} Accuracy vs.\ memory under the same $B_{\max}$; the full method (green) dominates ablated variants (blue).}
  \label{fig:ablation}
  \vspace{-16pt}
\end{figure}
\section{Conclusion}

In this work, we introduce AutoQRA. This framework jointly allocates quantization bit-width and LoRA rank for each layer. Conventional pipelines optimize these factors sequentially. That approach ignores the compensatory interaction between precision and adapter capacity. AutoQRA solves the discrete optimization problem using a coarse-to-fine strategy. The method combines multi-fidelity evolutionary search with trust-region Bayesian refinement. This design navigates the search space efficiently. Crucially, the search overhead is a one-time cost amortized over massive subsequent deployments. Extensive experiments validate our approach. AutoQRA consistently outperforms uniform 4-bit baselines on LLaMA and Qwen models. It matches full-precision performance while maintaining a lower average bit-width. We also identify a key compensatory pattern. The search algorithm automatically assigns higher ranks to layers with lower precision. This mechanism maximizes the utility of the limited memory budget. AutoQRA establishes a new standard for memory-efficient fine-tuning.

\section*{Impact Statement}

This paper presents work that advances the field of efficient LLM fine-tuning. Our primary goal is to enable high-performance model adaptation on consumer-grade hardware. AutoQRA significantly reduces the memory footprint required for training. This efficiency lowers the barrier for researchers and developers with limited computational budgets. It also contributes to environmental sustainability by decreasing the energy consumption of the adaptation process. We do not foresee specific negative societal consequences beyond the general risks associated with model compression and large language models.
\bibliography{main}

@misc{frankle2019lottery,
      title={The Lottery Ticket Hypothesis: Finding Sparse, Trainable Neural Networks}, 
      author={Jonathan Frankle and Michael Carbin},
      year={2019},
      eprint={1803.03635},
      archivePrefix={arXiv},
      primaryClass={cs.LG},
      url={https://arxiv.org/abs/1803.03635}, 
}

@misc{sta_lottery,
      title={Stabilizing the Lottery Ticket Hypothesis}, 
      author={Jonathan Frankle and Gintare Karolina Dziugaite and Daniel M. Roy and Michael Carbin},
      year={2020},
      eprint={1903.01611},
      archivePrefix={arXiv},
      primaryClass={cs.LG},
      url={https://arxiv.org/abs/1903.01611}, 
}

@article{deb2002fast,
  title={A fast and elitist multiobjective genetic algorithm: NSGA-II},
  author={Deb, Kalyanmoy and Pratap, Amrit and Agarwal, Sameer and Meyarivan, TAMT},
  journal={IEEE transactions on evolutionary computation},
  volume={6},
  number={2},
  pages={182--197},
  year={2002},
  publisher={IEEE}
}

@misc{QA-LoRA,
      title={QA-LoRA: Quantization-Aware Low-Rank Adaptation of Large Language Models}, 
      author={Yuhui Xu and Lingxi Xie and Xiaotao Gu and Xin Chen and Heng Chang and Hengheng Zhang and Zhengsu Chen and Xiaopeng Zhang and Qi Tian},
      year={2023},
      eprint={2309.14717},
      archivePrefix={arXiv},
      primaryClass={cs.LG},
      url={https://arxiv.org/abs/2309.14717}, 
}

@misc{huang2025slimllm,
      title={SliM-LLM: Salience-Driven Mixed-Precision Quantization for Large Language Models}, 
      author={Wei Huang and Haotong Qin and Yangdong Liu and Yawei Li and Qinshuo Liu and Xianglong Liu and Luca Benini and Michele Magno and Shiming Zhang and Xiaojuan Qi},
      year={2025},
      eprint={2405.14917},
      archivePrefix={arXiv},
      primaryClass={cs.LG},
      url={https://arxiv.org/abs/2405.14917}, 
}

@misc{amq,
      title={AMQ: Enabling AutoML for Mixed-precision Weight-Only Quantization of Large Language Models}, 
      author={Sangjun Lee and Seung-taek Woo and Jungyu Jin and Changhun Lee and Eunhyeok Park},
      year={2025},
      eprint={2509.12019},
      archivePrefix={arXiv},
      primaryClass={cs.LG},
      url={https://arxiv.org/abs/2509.12019}, 
}

@misc{wang2019haq,
      title={HAQ: Hardware-Aware Automated Quantization with Mixed Precision}, 
      author={Kuan Wang and Zhijian Liu and Yujun Lin and Ji Lin and Song Han},
      year={2019},
      eprint={1811.08886},
      archivePrefix={arXiv},
      primaryClass={cs.CV},
      url={https://arxiv.org/abs/1811.08886}, 
}

@misc{turbo,
      title={Scalable Global Optimization via Local Bayesian Optimization}, 
      author={David Eriksson and Michael Pearce and Jacob R Gardner and Ryan Turner and Matthias Poloczek},
      year={2020},
      eprint={1910.01739},
      archivePrefix={arXiv},
      primaryClass={cs.LG},
      url={https://arxiv.org/abs/1910.01739}, 
}

@misc{real2019regularized,
      title={Regularized Evolution for Image Classifier Architecture Search}, 
      author={Esteban Real and Alok Aggarwal and Yanping Huang and Quoc V Le},
      year={2019},
      eprint={1802.01548},
      archivePrefix={arXiv},
      primaryClass={cs.NE},
      url={https://arxiv.org/abs/1802.01548}, 
}

@article{zoph2016neural,
  title={Neural architecture search with reinforcement learning},
  author={Zoph, Barret and Le, Quoc V},
  journal={arXiv preprint arXiv:1611.01578},
  year={2016}
}

@inproceedings{frantar2023gptq,
  title={GPTQ: Accurate Post-Training Quantization for Generative Pre-trained Transformers},
  author={Frantar, Elias and Ashkboos, Sahar and Hoefler, Torsten and Alistarh, Dan},
  booktitle={The Eleventh International Conference on Learning Representations (ICLR)},
  year={2023}
}

@inproceedings{clark2019boolq,
  title={BoolQ: Exploring the Surprising Difficulty of Natural Yes/No Questions},
  author={Clark, Christopher and Lee, Kenton and Chang, Ming-Wei and Kwiatkowski, Tom and Collins, Michael and Toutanova, Kristina},
  booktitle={Proceedings of the 2019 Conference of the North American Chapter of the Association for Computational Linguistics: Human Language Technologies, Volume 1 (Long and Short Papers)},
  pages={2924--2936},
  year={2019}
}

@inproceedings{bisk2020piqa,
  title={Piqa: Reasoning about physical commonsense in natural language},
  author={Bisk, Yonatan and Zellers, Rowan and Gao, Jianfeng and Choi, Yejin and others},
  booktitle={Proceedings of the AAAI conference on artificial intelligence},
  volume={34},
  pages={7432--7439},
  year={2020}
}

@inproceedings{mihaylov2018can,
  title={Can a Suit of Armor Conduct Electricity? A New Dataset for Open Book Question Answering},
  author={Mihaylov, Todor and Clark, Peter and Khot, Tushar and Sabharwal, Ashish},
  booktitle={Proceedings of the 2018 Conference on Empirical Methods in Natural Language Processing},
  pages={2381--2391},
  year={2018}
}

@misc{zhao2025benchmarkingPTQ,
      title={Benchmarking Post-Training Quantization in LLMs: Comprehensive Taxonomy, Unified Evaluation, and Comparative Analysis}, 
      author={Jiaqi Zhao and Ming Wang and Miao Zhang and Yuzhang Shang and Xuebo Liu and Yaowei Wang and Min Zhang and Liqiang Nie},
      year={2025},
      eprint={2502.13178},
      archivePrefix={arXiv},
      primaryClass={cs.LG},
      url={https://arxiv.org/abs/2502.13178}, 
}

@article{lin2023awq,
  title={AWQ: Activation-aware weight quantization for LLM compression and acceleration},
  author={Lin, Ji and Tang, Jie and Tang, Haotao and Yang, Shuxin and Dang, Xiaoxia and Han, Song},
  journal={arXiv preprint arXiv:2306.00978},
  year={2023}
}

@article{clark2018think,
  author = {Clark, Peter and Cowhey, Isaac and Etzioni, Oren and Khot, Tushar and Sabharwal, Ashish and Schoenick, Carissa and Tafjord, Oyvind},
  title = {Think you have solved question answering? try arc, the ai2 reasoning challenge},
  journal = {arXiv preprint arXiv:1803.05457},
  year = {2018}
}

@article{dettmers2023qlora,
  author = {Dettmers, Tim and Pagnoni, Artidoro and Holtzman, Ari and Zettlemoyer, Luke},
  title = {Qlora: Efficient finetuning of quantized llms},
  journal = {arXiv preprint arXiv:2305.14314},
  year = {2023}
}

@article{guo2023lq,
  title={LQ-LoRA: Low-rank plus Quantized Matrix Decomposition for Efficient Language Model Finetuning},
  author={Guo, Han and Greengard, Philip and Xing, Eric and Kim, Yoon},
  journal={ICLR 2024},
  year={2023},
  publisher={ICLR 2024}
}

@inproceedings{hendrycks2021measuring,
  author = {Hendrycks, Dan and Burns, Collin and Basart, Steven and Zou, Andy and Mazeika, Mantas and Song, Dawn and Steinhardt, Jacob},
  title = {Measuring massive multitask language understanding},
  booktitle = {International Conference on Learning Representations},
  year = {2021},
  url = {https://openreview.net/forum?id=d7KBjmI3GmQ}
}

@inproceedings{hu2022lora,
  author = {Hu, Edward J. and Shen, Yelong and Wallis, Phillip and Allen-Zhu, Zeyuan and Li, Yuanzhi and Wang, Shean and Wang, Lu and Chen, Weizhu},
  title = {LoRA: Low-Rank Adaptation of Large Language Models},
  booktitle = {Proceedings of ICLR},
  year = {2022}
}

@article{sakaguchi2021winogrande,
  author = {Sakaguchi, Keisuke and Bras, Ronan Le and Bhagavatula, Chandra and Choi, Yejin},
  title = {Winogrande: An adversarial winograd schema challenge at scale},
  journal = {Communications of the ACM},
  volume = {64},
  number = {9},
  pages = {99--106},
  year = {2021}
}

@article{alpaca,
  author = {Taori, Rohan and Gulrajani, Ishaan and Zhang, Tianyi and Dubois, Yann and Li, Xuechen and Guestrin, Carlos and Liang, Percy and Hashimoto, Tatsunori B.},
  title = {Stanford alpaca: An instruction-following llama model},
  year = {2023},
  journal = {Stanford CRFM},
  url = {https://github.com/tatsu-lab/stanford_alpaca}
}

@inproceedings{zhou2025rankadaptor,
    title = "{R}ank{A}daptor: Hierarchical Rank Allocation for Efficient Fine-Tuning Pruned {LLM}s via Performance Model",
    author = "Zhou, Changhai  and
      Han, Shijie  and
      Yang, Lining  and
      Zhou, Yuhua  and
      Cheng, Xu  and
      Wang, Yibin  and
      Li, Hongguang",
    editor = "Chiruzzo, Luis  and
      Ritter, Alan  and
      Wang, Lu",
    booktitle = "Findings of the Association for Computational Linguistics: NAACL 2025",
    month = apr,
    year = "2025",
    address = "Albuquerque, New Mexico",
    publisher = "Association for Computational Linguistics",
    url = "https://aclanthology.org/2025.findings-naacl.321/",
    doi = "10.18653/v1/2025.findings-naacl.321",
    pages = "5781--5795",
    ISBN = "979-8-89176-195-7"
}

@misc{qwen2025qwen25technicalreport,
      title={Qwen2.5 Technical Report}, 
      author={Qwen and : and An Yang and Baosong Yang and Beichen Zhang and Binyuan Hui, et al.},
      year={2025},
      eprint={2412.15115},
      archivePrefix={arXiv},
      primaryClass={cs.CL},
      url={https://arxiv.org/abs/2412.15115}, 
}

@article{chang2024survey,
  title={A survey on evaluation of large language models},
  author={Chang, Yupeng and Wang, Xu and Wang, Jindong and Wu, Yuan and Yang, Linyi and Zhu, Kaijie and Chen, Hao and Yi, Xiaoyuan and Wang, Cunxiang and Wang, Yidong and others},
  journal={ACM Transactions on Intelligent Systems and Technology},
  volume={15},
  number={3},
  pages={1--45},
  year={2024},
  publisher={ACM New York, NY}
}

@article{makridakis2023large,
  title={Large language models: Their success and impact},
  author={Makridakis, Spyros and Petropoulos, Fotios and Kang, Yanfei},
  journal={Forecasting},
  volume={5},
  number={3},
  pages={536--549},
  year={2023},
  publisher={MDPI}
}

@article{raiaan2024review,
  title={A review on large Language Models: Architectures, applications, taxonomies, open issues and challenges},
  author={Raiaan, Mohaimenul Azam Khan and Mukta, Md Saddam Hossain and Fatema, Kaniz and Fahad, Nur Mohammad and Sakib, Sadman and Mim, Most Marufatul Jannat and Ahmad, Jubaer and Ali, Mohammed Eunus and Azam, Sami},
  journal={IEEE Access},
  year={2024},
  publisher={IEEE}
}

@inproceedings{zellers2019hellaswag,
  author = {Zellers, Rowan and Holtzman, Ari and Bisk, Yonatan and Farhadi, Ali and Choi, Yejin},
  title = {Hellaswag: Can a machine really finish your sentence?},
  booktitle = {Proceedings of the 57th Annual Meeting of the Association for Computational Linguistics},
  pages = {4791--4800},
  year = {2019}
}

@inproceedings{
guo2024lqlora,
title={{LQ}-Lo{RA}: Low-rank plus Quantized Matrix Decomposition for Efficient Language Model Finetuning},
author={Han Guo and Philip Greengard and Eric Xing and Yoon Kim},
booktitle={The Twelfth International Conference on Learning Representations},
year={2024},
url={https://openreview.net/forum?id=xw29VvOMmU}
}

@article{zhang2023adalora,
  title={AdaLoRA: Adaptive budget allocation for parameter-efficient fine-tuning},
  author={Zhang, Qingru and Chen, Minshuo and Bukharin, Alexander and Karampatziakis, Nikos and He, Pengcheng and Cheng, Yu and Chen, Weizhu and Zhao, Tuo},
  journal={arXiv preprint arXiv:2303.10512},
  year={2023}
}

@misc{li2023loftq,
      title={LoftQ: LoRA-Fine-Tuning-Aware Quantization for Large Language Models}, 
      author={Yixiao Li and Yifan Yu and Chen Liang and Pengcheng He and Nikos Karampatziakis and Weizhu Chen and Tuo Zhao},
      year={2023},
      eprint={2310.08659},
      archivePrefix={arXiv},
      primaryClass={cs.CL},
      url={https://arxiv.org/abs/2310.08659}, 
}

@misc{grattafiori2024llama3herdmodels,
      title={The Llama 3 Herd of Models}, 
      author={Aaron Grattafiori and Abhimanyu Dubey and Abhinav Jauhri and Abhinav Pandey, et al.},
      year={2024},
      eprint={2407.21783},
      archivePrefix={arXiv},
      primaryClass={cs.AI},
      url={https://arxiv.org/abs/2407.21783}, 
}

@misc{li2018hyperband,
      title={Hyperband: A Novel Bandit-Based Approach to Hyperparameter Optimization}, 
      author={Lisha Li and Kevin Jamieson and Giulia DeSalvo and Afshin Rostamizadeh and Ameet Talwalkar},
      year={2018},
      eprint={1603.06560},
      archivePrefix={arXiv},
      primaryClass={cs.LG},
      url={https://arxiv.org/abs/1603.06560}, 
}

@misc{falkner2018bohb,
      title={BOHB: Robust and Efficient Hyperparameter Optimization at Scale}, 
      author={Stefan Falkner and Aaron Klein and Frank Hutter},
      year={2018},
      eprint={1807.01774},
      archivePrefix={arXiv},
      primaryClass={cs.LG},
      url={https://arxiv.org/abs/1807.01774}, 
}

@misc{ru2020cocabo,
      title={Bayesian Optimisation over Multiple Continuous and Categorical Inputs}, 
      author={Binxin Ru and Ahsan S. Alvi and Vu Nguyen and Michael A. Osborne and Stephen J Roberts},
      year={2020},
      eprint={1906.08878},
      archivePrefix={arXiv},
      primaryClass={stat.ML},
      url={https://arxiv.org/abs/1906.08878}, 
}

@article{jones1998ego,
  title   = {Efficient Global Optimization of Expensive Black-Box Functions},
  author  = {Jones, Donald R. and Schonlau, Matthias and Welch, William J.},
  journal = {Journal of Global Optimization},
  year    = {1998},
  volume  = {13},
  number  = {4},
  pages   = {455--492},
  doi     = {10.1023/A:1008306431147},
  url     = {https://doi.org/10.1023/A:1008306431147}
}

@article{zitzler1999hypervolume,
  title   = {Multiobjective evolutionary algorithms: A comparative case study and the strength Pareto approach},
  author  = {Zitzler, Eckart and Thiele, Lothar},
  journal = {IEEE Transactions on Evolutionary Computation},
  year    = {1999},
  volume  = {3},
  number  = {4},
  pages   = {257--271},
  doi     = {10.1109/4235.797969}
}
\bibliographystyle{icml2026}

\clearpage
\appendix
\setcounter{page}{1}

\section{Orthogonal Sensitivity and Compensatory Potential.}
\label{app:Orthogonal}
\begin{figure}[t]
  \centering
  \includegraphics[width=\linewidth]{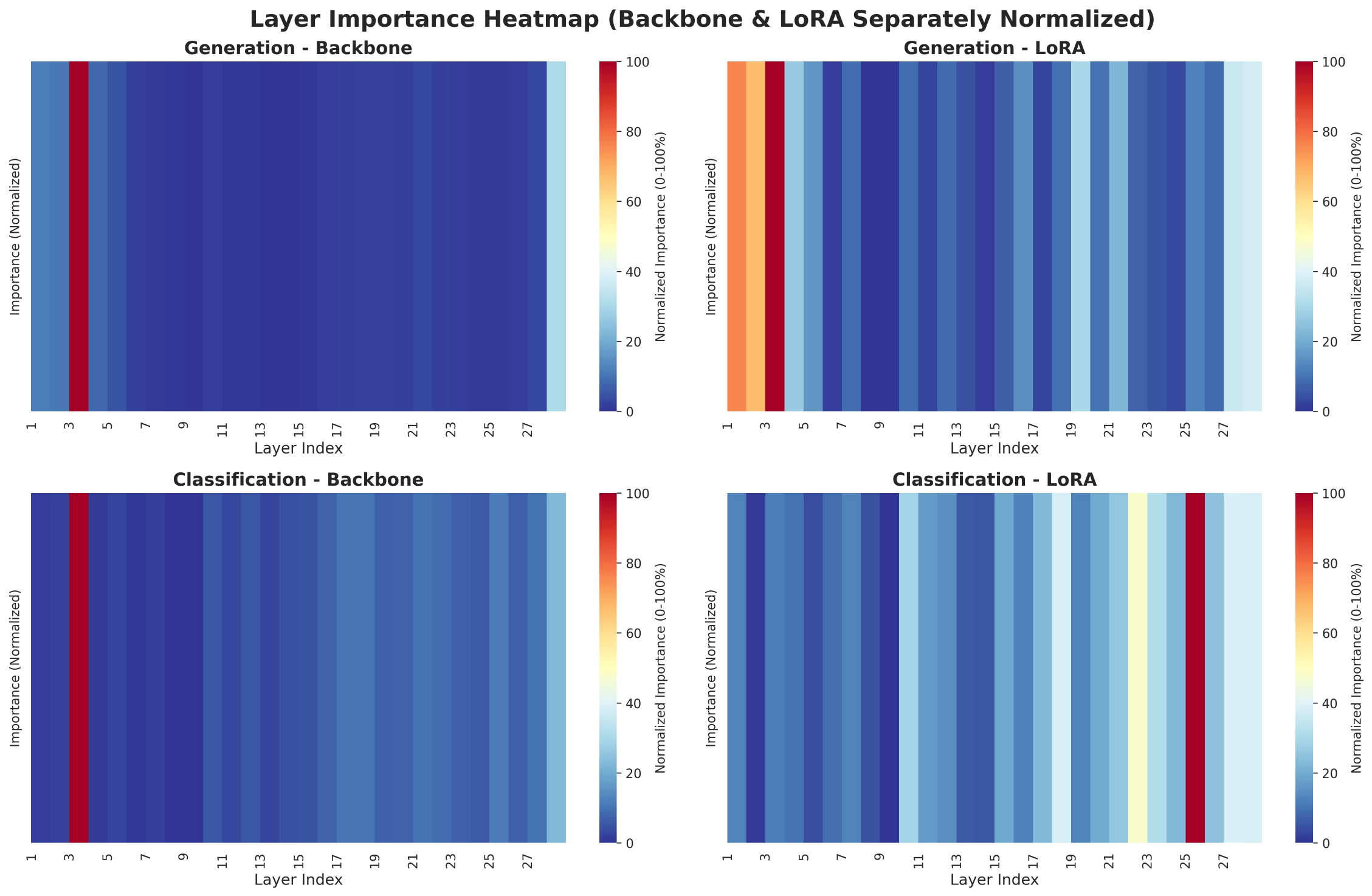}
  \vspace{-15pt}
  \caption{\textbf{Orthogonal sensitivity profiles in Qwen3-1.7B.} Normalized importance scores for quantization (backbone weight sensitivity, left) and adaptation (LoRA rank sensitivity, right). Crucially, these distributions diverge: layers requiring high precision to suppress quantization noise do not necessarily demand high-rank adapters for downstream tasks. This \emph{sensitivity mismatch} necessitates joint allocation, as independent optimization fails to capture these compensatory trade-offs.}
  \label{fig:important}
  \vspace{-10pt}
\end{figure}

While layer-wise heterogeneity is well-documented, the primary rationale for joint optimization lies in the \emph{orthogonality} of sensitivity profiles. As illustrated in Figure~\ref{fig:important}, layers highly sensitive to quantization (requiring high bit-width) often differ from those requiring high adaptation capacity (requiring high rank).
Conventional sequential pipelines overlook this nuance: a quantization-first approach might conservatively assign 4-bit precision to a sensitive layer, ignoring that a high-rank adapter could effectively compensate for the noise induced by aggressive 2-bit quantization. Conversely, independent optimization might allocate high ranks to layers robust to quantization but contributing little to task adaptation. Therefore, optimal allocation requires capturing this \emph{compensatory interplay}: trading precision for learnability based on layer-specific joint sensitivity.

\section{Additional Details for Phase I}
\label{app:phase1_details}

\paragraph{Exact memory decomposition.}
For feasibility checks we use the exact memory accounting implemented in our code, including quantization metadata such as scales and zero-points.
For readability, the footprint can be decomposed as
\begin{equation}
\label{eq:app_mem_decomp}
M(C)=\sum_{\ell=1}^L \Big(m^{W}_\ell(q_\ell) + m^{A}_\ell(r_\ell) + m^{\mathrm{meta}}_\ell(q_\ell)\Big),
\end{equation}
where $m^{W}_\ell$ is the quantized backbone storage, $m^{A}_\ell$ is the adapter storage, and $m^{\mathrm{meta}}_\ell$ captures quantization metadata.
Letting $N(W_\ell)$ denote the number of backbone parameters at layer $\ell$, we write
\begin{equation}
\label{eq:app_mem_terms}
m^{W}_\ell(q_\ell)=\frac{N(W_\ell)\,q_\ell}{8},
\qquad
m^{A}_\ell(r_\ell)=\frac{N(A_\ell,r_\ell)\,p_r}{8},
\end{equation}
with $p_r=16$ for FP16 adapters.
For LoRA applied to a set of linear maps $\mathcal{S}_\ell$ in layer $\ell$, the adapter parameter count is
\begin{equation}
\label{eq:app_lora_param_count}
N(A_\ell,r_\ell)=\sum_{W\in\mathcal{S}_\ell} r_\ell\big(d_{\text{in}}(W)+d_{\text{out}}(W)\big),
\end{equation}
following the standard LoRA parameterization \citep{hu2022lora}.

\begin{table*}[t]
\centering
\scriptsize
\setlength{\tabcolsep}{2.6pt}
\renewcommand{\arraystretch}{1.05}
\caption{\textbf{Detailed Task-wise accuracy.} Bold indicates best result per backbone.}
\label{tab:main_four_models}
\resizebox{\linewidth}{!}{
\begin{tabular}{llccc ccccccccc}
\toprule
\textbf{Model} & \textbf{Method} & \textbf{Bit} & \textbf{Rank} & \textbf{Mem} &
\textbf{Avg} & \textbf{ARC-C} & \textbf{ARC-E} & \textbf{BoolQ} & \textbf{GSM8K} &
\textbf{HellaS} & \textbf{OBQA} & \textbf{PIQA} & \textbf{WinoG} \\
\midrule
 \multirow{9}{*}{\textbf{LLaMA-3.1-8B}} & LoRA & 16.00 & 16.00 & 20.50 & 69.94 & 56.14 & 83.88 & 83.37 & \textbf{54.06} & 79.44 & \textbf{45.60} & 82.10 & \textbf{74.90} \\
  & QLoRA & 4.00 & 16.00 & 15.22 & 67.45 & 54.15 & 82.20 & 81.95 & 44.15 & 78.50 & 43.60 & 81.40 & 73.64 \\
  & AdaLoRA & 4.00 & 15.84 & 14.92 & 66.36 & 52.40 & 81.65 & 81.50 & 42.80 & 78.10 & 40.80 & 81.10 & 72.55 \\
  & LoftQ & 4.00 & 16.00 & 15.13 & 68.65 & 54.80 & 82.65 & 82.15 & 51.10 & 78.65 & 45.00 & 81.45 & 73.40 \\
  & LQ-LoRA & 3.78 & 16.00 & 20.12 & 67.82 & 54.75 & 82.45 & 82.25 & 44.95 & 79.10 & 44.20 & 81.60 & 73.25 \\
  & AMQ+L & 3.88 & 16.00 & 14.45 & 67.75 & 54.55 & 82.60 & 82.40 & 44.50 & 78.90 & 44.00 & 81.42 & 73.61 \\
  & AMQ+AL & 3.93 & 10.18 & 14.40 & 67.63 & 54.45 & 82.35 & 82.15 & 44.25 & 78.60 & 43.80 & 81.76 & 73.69 \\
  & \textbf{AutoQRA} ($\le$4b) & 3.75 & 10.50 & 13.08 & 69.83 & 56.12 & 84.20 & 83.35 & 53.40 & 79.50 & 45.60 & 82.40 & 74.10 \\
  & \textbf{AutoQRA (Opt)} & 5.25 & 12.25 & 17.32 & \textbf{70.45} & \textbf{57.85} & \textbf{84.90} & \textbf{83.86} & 53.90 & \textbf{79.95} & 45.20 & \textbf{83.10} & 74.80 \\
\midrule
 \multirow{9}{*}{\textbf{LLaMA-3.2-3B}} & LoRA & 16.00 & 16.00 & 10.40 & 65.40 & 48.38 & \textbf{79.78} & \textbf{81.10} & 47.31 & 75.75 & 42.20 & 78.62 & 70.09 \\
  & QLoRA & 4.00 & 16.00 & 8.90 & 64.43 & 47.18 & 77.53 & 77.61 & 46.22 & 74.79 & 42.40 & 78.73 & 70.96 \\
  & AdaLoRA & 4.00 & 15.73 & 8.60 & 61.88 & 46.42 & 74.79 & 74.68 & 36.50 & 73.92 & 41.20 & 77.97 & 69.53 \\
  & LoftQ & 4.00 & 16.00 & 8.83 & 64.91 & 47.80 & 77.90 & 77.95 & 47.50 & 75.10 & 42.80 & 79.00 & 71.20 \\
  & LQ-LoRA & 3.85 & 16.00 & 7.12 & 63.63 & \textbf{51.45} & 79.20 & 79.05 & 36.40 & 75.25 & 40.40 & 78.20 & 69.10 \\
  & AMQ+L & 3.91 & 16.00 & 7.22 & 63.31 & 46.90 &	75.90 & 76.80 & 42.10&73.90&41.50&78.90&70.50 \\
  & AMQ+AL & 3.96 & 9.68 & 7.25 & 63.38 & 47.30 & 79.25 & 79.40 & 36.85 & 75.50 & 40.80 & 78.50 & 69.45 \\
  & \textbf{AutoQRA} ($\le$4b) & 3.64 & 11.14 & 6.72 & 65.58 & 47.97 & 78.96 & 78.59 & 48.93 & 76.97 & 42.80 & 79.15 & 71.20 \\
  & \textbf{AutoQRA (Opt)} & 5.14 & 12.57 & 8.41 & \textbf{66.16} & 48.00 & 79.60 & 79.70 & \textbf{49.72} & \textbf{77.29} & \textbf{43.80} & \textbf{79.40} & \textbf{71.80} \\
\midrule
 \multirow{9}{*}{\textbf{Qwen-2.5-7B}} & LoRA & 16.00 & 16.00 & 20.20 & 71.33 & 53.16 & 79.21 & 84.43 & 75.24 & 77.87 & 46.00 & 81.66 & 73.04 \\
  & QLoRA & 4.00 & 16.00 & 15.60 & 69.01 & 54.27 & 79.04 & 83.73 & 73.50 & 78.04 & 43.80 & 76.39 & 63.30 \\
  & AdaLoRA & 4.00 & 15.65 & 15.40 & 69.05 & 50.85 & 80.40 & 80.00 & 69.80 & 77.00 & 42.00 & 80.45 & 71.90 \\
  & LoftQ & 4.00 & 16.00 & 15.33 & 69.35 & 53.92 & 78.83 & 84.25 & 75.10 & 78.17 & 43.00 & 77.37 & 64.17 \\
  & LQ-LoRA & 3.85 & 16.00 & 18.52 & 66.91 & 53.70 & 81.70 & 81.25 & 44.00 & 77.80 & 43.80 & 80.50 & 72.55 \\
  & AMQ+L & 3.91 & 16.00 & 14.90 & 70.86 & 53.10 & 81.40 & 82.10 & 74.80 & 77.90 & 43.40 & 81.10 & 73.10 \\
  & AMQ+AL & 3.96 & 10.18 & 14.70 & 70.66 & 53.80 & 81.35 & 81.15 & 74.20 & 77.55 & 43.20 & 81.00 & 73.05 \\
  & \textbf{AutoQRA} ($\le$4b) & 3.71 & 8.57 & 10.57 & 71.35 & 54.30 & 81.26 & 83.74 & 74.86 & 78.46 & 43.60 & 81.30 & 73.30 \\
  & \textbf{AutoQRA (Opt)} & 5.36 & 13.18 & 12.70 & \textbf{73.19} & \textbf{55.75} & \textbf{82.93} & \textbf{85.86} & \textbf{76.28} & \textbf{79.33} & \textbf{46.80} & \textbf{83.43} & \textbf{75.12} \\
\midrule
 \multirow{9}{*}{\textbf{Qwen-2.5-3B}} & LoRA & 16.00 & 16.00 & 10.84 & 65.53 & 48.20 & 75.10 & 80.00 & 64.00 & 75.50 & 35.00 & 77.00 & 69.40 \\
  & QLoRA & 4.00 & 16.00 & 8.16 & 62.89 & 43.60 & 72.69 & 78.65 & 64.90 & 73.43 & 34.40 & 72.52 & 62.90 \\
  & AdaLoRA & 4.00 & 15.92 & 8.70 & 62.51 & 42.50 & 75.57 & 76.09 & 68.30 & 70.70 & 28.20 & 74.62 & 64.10 \\
  & LoftQ & 4.00 & 16.00 & 8.21 & 62.71 & 44.62 & 73.27 & 80.37 & 63.38 & 73.75 & 30.00 & 72.96 & 63.30 \\
  & LQ-LoRA & 3.63 & 16.00 & 7.05 & 62.86 & \textbf{50.50} & 78.30 & 78.30 & 35.25 & 74.60 & \textbf{40.00} & \textbf{77.40} & 68.50 \\
  & AMQ+L & 4.00 & 16.00 & 8.17 & 63.30 & 44.80 & 74.10 & 78.20 & 64.10 & 73.90 & 34.60 & 73.10 & 63.60 \\
  & AMQ+AL & 3.80 & 15.84 & 8.65 & 64.88 & 45.60 & 76.54 & 77.90 & 72.50 & 71.80 & 31.40 & 76.13 & 67.20 \\
  & \textbf{AutoQRA} ($\le$4b) & 3.72 & 9.78 & 6.45 & 66.33 & 46.35 & 77.23 & 80.12 & 72.88 & 75.19 & 35.20 & 75.60 & 68.09 \\
  & \textbf{AutoQRA (Opt)} & 5.22 & 12.00 & 8.31 & \textbf{68.05} & 47.31 & \textbf{79.32} & \textbf{82.21} & \textbf{74.00} & \textbf{77.52} & 36.80 & 76.86 & \textbf{70.40} \\
\bottomrule
\end{tabular}}
\end{table*}

\paragraph{Importance signals for warm start and proposals.}
We define two layerwise signals used only for warm start and proposal distributions in Phase~I (and for feasibility repair), not as substitutes for the final fine-tuning metric.
For bit-width proposals, we use a gradient-weighted quantization residual on a small calibration set.
Let $\mathcal{S}^W_\ell$ denote backbone matrices in layer $\ell$, let $q_{\min}=\min\mathcal{Q}$, and let $Q_{q_{\min}}(\cdot)$ denote quantization at $q_{\min}$.
With a diagonal Fisher proxy $\widehat{F}_W\approx \mathbb{E}\!\left[\left(\frac{\partial \mathcal{L}}{\partial W}\right)^{\odot 2}\right]$, we define
\begin{equation}
\label{eq:app_iq_def}
I_q(\ell)=\sum_{W\in\mathcal{S}^W_\ell}
\Big\langle \widehat{F}_W,\; \big(W-Q_{q_{\min}}(W)\big)^{\odot 2}\Big\rangle .
\end{equation}
For rank proposals, we run a short probe fine-tuning for $K$ steps and collect the averaged gradient matrix for each LoRA-targeted projection $W\in\mathcal{S}_\ell$:
\begin{equation}
\label{eq:app_grad_probe}
\overline{G}_W=\frac{1}{K}\sum_{k=1}^K \frac{\partial \mathcal{L}_k}{\partial W}.
\end{equation}
Let $\{\sigma_j(\overline{G}_W)\}$ be the singular values of $\overline{G}_W$. We define the rank signal by the leading spectral energy:
\begin{equation}
\label{eq:app_ir_def}
I_r(\ell)=\sum_{W\in\mathcal{S}_\ell}\sum_{j=1}^{J_0}\sigma_j(\overline{G}_W)^2,
\end{equation}
where $J_0$ is a small cutoff (e.g., $J_0=8$) to emphasize dominant update directions.
In Phase~I we min--max normalize these scores across layers and use them for warm-start initialization, importance-guided proposals, and feasibility repair.

\section{Experimental Details}
\label{app:details}

\subsection{Phase I multi-fidelity step allocation}
We use a Hyperband/SHA-style ladder with step counts $0<T_1<\cdots<T_S$ and pruning factor $\eta$.
In our default setting, we evaluate $N_{\mathrm{LF}}=25$ configurations at the lowest fidelity $T_1$.
Candidates are ranked for promotion using the surrogate screening model $\Phi_s$ (Eq.~\eqref{eq:screen_loss}) when enough paired data are available;
otherwise we fall back to the measured $P(C;T_s)$.
After successive halving, $N_{\mathrm{HF}}=3$ configurations are promoted to the highest-fidelity step count $T_S$ (high-fidelity, HF),
\emph{continuing training from their checkpoints} rather than restarting.
All feasibility checks use the exact memory accounting in implementation.

\subsection{Phase I termination}
\label{app:Hypervolume}
We terminate Phase~I using the hypervolume-based criterion in Eq.~\eqref{eq:hv_stop}:
if the relative improvement of dominated hypervolume stays below $\epsilon_{\mathrm{hv}}$ for $\Delta$ consecutive generations,
we stop and return the measured feasible non-dominated set $\mathcal{P}$.

After each generation, we update the population using NSGA-II with constrained domination:
feasible candidates dominate infeasible ones; among infeasible candidates, lower constraint violation is preferred \citep{deb2002fast}.
To quantify progress of the feasible non-dominated set, we monitor the dominated hypervolume in the bi-objective space
(maximize $P$, minimize $M$). We map each configuration to a minimization vector
$\mathbf{f}(C)=\big(-P(C;T_S),\, M(C)\big)$ and define the hypervolume of a set $\mathcal{A}$ w.r.t.\ a reference pint $\mathbf{r}$ as
\begin{equation}
\label{eq:hv_def}
\begin{split}
\mathrm{HV}(\mathcal{A};\mathbf{r})
&=\lambda\!\left(\bigcup_{\mathbf{a}\in \mathbf{f}(\mathcal{A})} [\mathbf{a},\mathbf{r}]\right), \\
[\mathbf{a},\mathbf{r}]
&=\{\mathbf{z}:\ a_k\le z_k\le r_k,\ \forall k\},
\end{split}
\end{equation}
where $\lambda(\cdot)$ denotes the Lebesgue measure and $\mathbf{f}(\mathcal{A})=\{\mathbf{f}(C):C\in\mathcal{A}\}$.
We use the relative hypervolume improvement
\begin{equation}
\label{eq:hv_stop}
\Delta\mathrm{HV}_g
=\frac{\mathrm{HV}(\mathcal{P}_g;\mathbf{r})-\mathrm{HV}(\mathcal{P}_{g-1};\mathbf{r})}
{\mathrm{HV}(\mathcal{P}_{g-1};\mathbf{r})+\epsilon_{\mathrm{hv}}^{\mathrm{den}}},
\end{equation}
and stop Phase~I if $\Delta\mathrm{HV}_g<\epsilon_{\mathrm{hv}}$ for $\Delta$ consecutive generations.
Here $\epsilon_{\mathrm{hv}}^{\mathrm{den}}$ is a small constant to avoid numerical issues when the denominator is tiny.
We return the final non-dominated feasible set $\mathcal{P}$ for Phase~II.

\subsection{Phase II TuRBO high-fidelity evaluation}
Phase~II initializes $J$ trust regions from the $N_{\mathrm{HF}}$ measured points returned by Phase~I.
At iteration $t$, we form a discrete candidate pool $\Omega_t$ (Eqs.~\eqref{eq:phase2_ball}--\eqref{eq:phase2_tr})
and compute $\mathrm{EI}_t(C)$ for all $C\in\Omega_t$ using the GP posterior.
Only the maximizer $C_{t+1}$ in Eq.~\eqref{eq:phase2_next} is evaluated at the highest-fidelity step count $T_S$.
We set a hard cap of $N_{\max}$ Phase~II iterations.
Optionally, we also cap the number of accepted HF evaluations per trust region by $N_{\max}^{\mathrm{TR}}$ (default $N_{\max}^{\mathrm{TR}}=5$),
so the worst-case additional HF evaluations in Phase~II is at most $J\cdot N_{\max}^{\mathrm{TR}}$, but typically much smaller due to early stopping.

\subsection{Phase II early stopping}
We stop Phase~II when either (i) $\max_{C\in\Omega_t}\mathrm{EI}_t(C)<\epsilon_{\mathrm{ei}}$ (Eq.~\eqref{eq:phase2_stop}),
or (ii) the hard cap $N_{\max}$ is reached.

\begin{figure}[t]
  \centering
  \includegraphics[width=\linewidth]{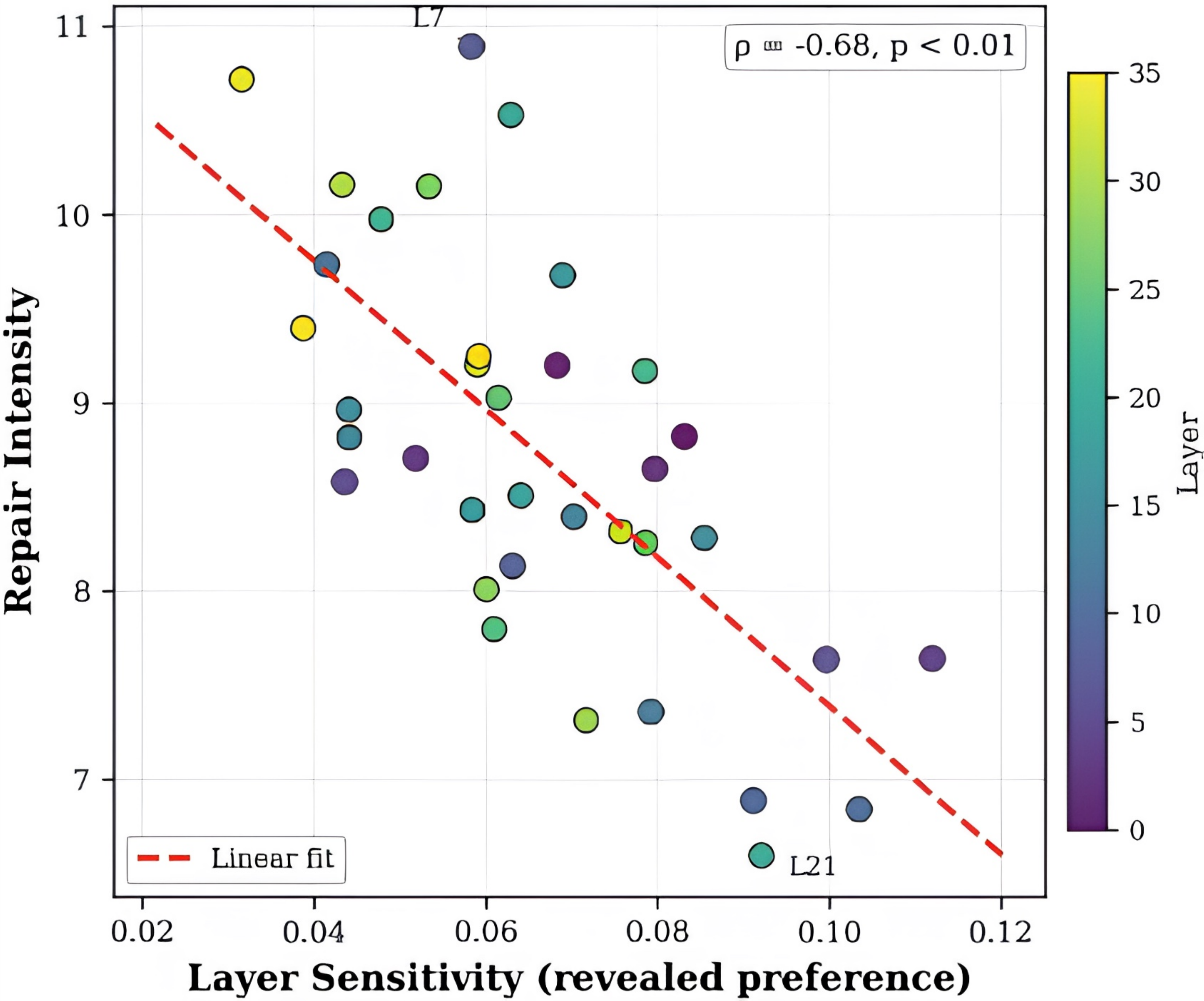}
  \vspace{-0.6em}
  \caption{\textbf{Repair concentrates on robust layers.}
Layer-wise repair intensity versus layer sensitivity (revealed preference).
Each point is a layer (color indicates layer index); the dashed line is a linear fit.
Repair intensity is strongly negatively correlated with sensitivity ($\rho=-0.68$, $p<0.01$),
showing that $\Repair(\cdot)$ preferentially applies downgrades to less sensitive layers to satisfy $M(C)\le B_{\max}$.}
  \label{fig:repair_summary}
  \vspace{-0.8em}
\end{figure}

\begin{figure}[t]
  \centering
  \includegraphics[width=0.95\linewidth]{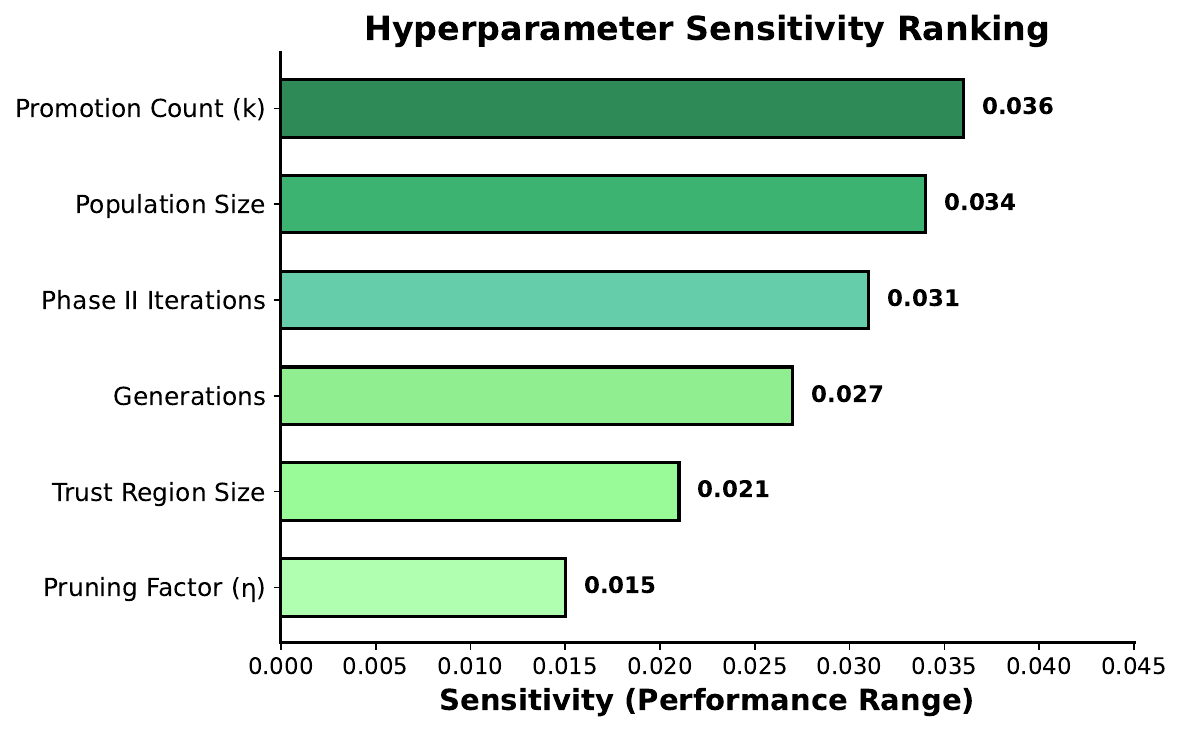}
  \vspace{-0.4em}
  \caption{\textbf{Hyperparameter sensitivity ranking.}
  Sensitivity is measured as $S(h)$ in Eq.~\eqref{eq:hparam_sensitivity} (max--min performance across the sweep of $h$);
  larger values indicate higher sensitivity.}
  \label{fig:hparam_rank}
  \vspace{-0.8em}
\end{figure}

\section{Task-wise Accuracy Breakdown and Discussion}
\label{app:taskwise_discussion}

Table~\ref{tab:main_four_models} breaks down accuracy by task for all four backbones.
Averaged scores can hide where methods actually win or lose, so here we focus on the “shape” of the gains: which tasks are stable under aggressive compression, and which ones are the first to fall apart.			

\paragraph{First takeaway: the uniform 4-bit baselines fail in very specific ways, and AutoQRA mostly fixes those failures.}
Across backbones, QLoRA/AdaLoRA/LoftQ are often fairly close on the easier or more pattern-heavy tasks, but they can drop sharply on a few tasks that are more brittle to quantization noise.
The cleanest example is \textbf{Qwen-2.5-7B}: QLoRA collapses on \textbf{PIQA} (76.39) and especially \textbf{WinoGrande} (63.30), even though its average is not catastrophically low.
AutoQRA ($\le$4b) brings these back to 81.30 and 73.30, essentially recovering the FP16 behavior (81.66 / 73.04) while staying in the low-precision regime.
So the main story is not that AutoQRA gives tiny uniform gains everywhere---it prevents the “one or two tasks tank” pattern that you get with a rigid allocation.

\paragraph{LLaMA-3.1-8B: AutoQRA ($\le$4b) is basically “FP16-like” on most tasks, with the biggest win on GSM8K.}
On this backbone, the uniform 4-bit methods lose most noticeably on \textbf{GSM8K} (e.g., QLoRA 44.15 vs.\ LoRA 54.06), while other tasks move less.
AutoQRA ($\le$4b) almost fully closes that gap (53.40), and it does so without trading away the rest: ARC-E (84.20 vs.\ 83.88), HellaSwag (79.50 vs.\ 79.44), and PIQA (82.40 vs.\ 82.10) are all on par with or slightly above FP16 LoRA.
Interestingly, \textbf{OpenBookQA} is basically flat across strong methods (AutoQRA ($\le$4b) 45.60, LoRA 45.60), which suggests OBQA is not where the low-precision decisions are biting most for this model.

\begin{figure*}[ht]
  \centering
  \includegraphics[width=\linewidth]{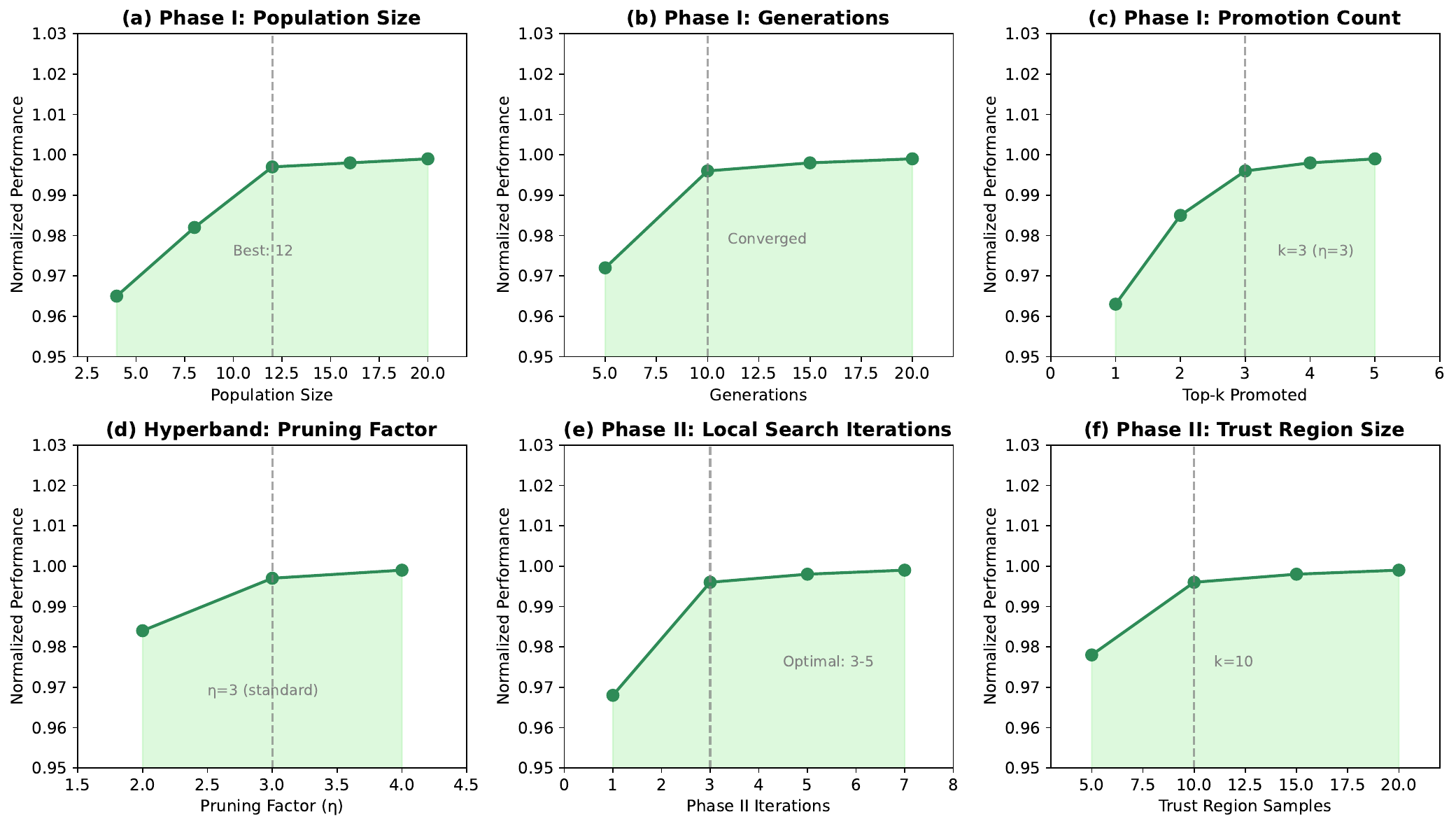}
  \vspace{-0.4em}
  \caption{\textbf{One-factor hyperparameter sweeps.}
  Each panel varies one hyperparameter while keeping all others fixed.
  The dashed vertical line denotes the default setting used in all experiments; curves show mean performance.}
  \label{fig:hparam_sweeps}
  \vspace{-0.8em}
\end{figure*}

\paragraph{LLaMA-3.2-3B: the gains are smaller and less uniform, but the “compensation” pattern still shows up where it matters.}
For the smaller LLaMA, AutoQRA ($\le$4b) still improves \textbf{GSM8K} (48.93 vs.\ QLoRA 46.22 / LoRA 47.31), \textbf{HellaSwag} (76.97 vs.\ 74.79 / 75.75), and \textbf{WinoGrande} (71.20 vs.\ 70.96 / 70.09).
At the same time, \textbf{BoolQ} drops for most low-precision methods and does not fully recover (AutoQRA (Opt) 79.70 vs.\ LoRA 81.10).
This is a useful sanity check: once the backbone is small, you cannot expect “free lunch” behavior.
Joint allocation helps, but some tasks remain sensitive to the exact capacity/precision trade, and the optimizer will sometimes take a small hit on one task to buy larger gains elsewhere (because the search objective is the post-finetuning average).

\paragraph{Qwen-2.5-3B: AutoQRA’s gains look like a re-prioritization toward multi-step reasoning.}
This backbone shows one of the clearest shifts on \textbf{GSM8K}: AutoQRA ($\le$4b) reaches 72.88, substantially above LoRA (64.00) and well above uniform 4-bit baselines.
At the same time, improvements on some other tasks are more modest (e.g., ARC-E 77.23 vs.\ 75.10, BoolQ 80.12 vs.\ 80.00), and a couple of tasks can move slightly in the opposite direction (e.g., PIQA 75.60 vs.\ 77.00).
The overall picture matches the intended behavior of AutoQRA: it is not “just quantize less everywhere,” but rather it reallocates capacity so that the tasks that are most disrupted by quantization noise are the ones that get the most help from rank.

\paragraph{AutoQRA ($\le$4b) vs.\ AutoQRA (Opt): what the extra precision buys.}
The (Opt) setting consistently raises the ceiling, but the pattern of gains matters.
On \textbf{Qwen-2.5-7B}, the jump from AutoQRA ($\le$4b) to (Opt) is broad and clean (Avg 71.35 $\rightarrow$ 73.19), and it also improves the previously brittle tasks further (e.g., WinoGrande 73.30 $\rightarrow$ 75.12, PIQA 81.30 $\rightarrow$ 83.43).
On \textbf{LLaMA-3.1-8B}, the (Opt) gains are more concentrated (ARC-C 56.12 $\rightarrow$ 57.85, ARC-E 84.20 $\rightarrow$ 84.90), while most other tasks are already near saturation under the $\le$4b regime.
So in practice, (Opt) mostly helps when you have headroom to push the “hard reasoning” tasks above an already-strong baseline, whereas $\le$4b is doing the heavy lifting for robustness under strict compression.

\paragraph{Bottom line.}
The per-task breakdown supports the core motivation: mixed bit-width choices that look acceptable in aggregate can still be fragile on specific tasks, and the joint bit--rank search largely removes these fragilities.
Where the backbone has enough capacity (8B/7B), AutoQRA ($\le$4b) behaves like a drop-in replacement for FP16 LoRA on most tasks while avoiding the sharp failure modes of uniform 4-bit methods.
Where the backbone is small (3B), the trade-offs become more visible, but the improvements still show up consistently on the tasks that are typically hardest to preserve under quantization.

\section{Additional Ablations}
\label{app:additional_ablation}

\subsection{Feasibility Repair Analysis}
\label{app:repair}

Figure~\ref{fig:repair_summary} gives a layer-level view of what $\Repair(\cdot)$ is \emph{actually} doing when it projects an infeasible configuration back into the feasible set.
For each layer, we plot its sensitivity score (x-axis; a revealed-preference measure) against the corresponding \emph{repair intensity} (y-axis; how strongly that layer tends to be hit by discrete downgrades during repair).
The pattern is very clear: repair intensity is strongly \emph{negatively} correlated with sensitivity ($\rho=-0.68$, $p<0.01$).
In other words, layers that are more sensitive are systematically protected, while the ``damage'' is concentrated on layers that are relatively robust.

This is exactly the behavior encouraged by Eq.~\eqref{eq:repair_rule}.
When $M(C)>B_{\max}$, $\Repair$ looks for the downgrade that removes the most memory for the least expected loss, i.e., small sensitivity-per-saved-memory.
The scatter plot shows that this heuristic translates into a consistent, global preference rather than a few hand-picked cases: across depth (color-coded layer index), repair pressure shifts toward low-sensitivity layers, effectively using them as a buffer to satisfy the hard constraint without repeatedly harming fragile layers.
A few outliers (e.g., the labeled layers) are expected in a discrete ladder setting: if a particular layer offers an unusually favorable memory drop for a single step on either the bit or rank ladder, it can be selected more/less often even when its sensitivity is not extreme.
Overall, the takeaway is that $\Repair(\cdot)$ is not just ``making things feasible''---it enforces feasibility in a way that preserves trainability by avoiding heavy edits on sensitive layers.

\subsection{Search-Protocol Hyperparameter Sensitivity}
\label{app:hparam_sens}

AutoQRA contains several discrete hyperparameters that control (i) Phase~I multi-fidelity evolutionary search
(e.g., population size, number of generations, promotion rule, and pruning factor $\eta$),
and (ii) Phase~II local refinement (e.g., EI iteration budget, scalarization weight $\alpha$, and the neighborhood size in discrete trust regions).
We assess robustness via a one-factor-at-a-time sweep: for each hyperparameter $h$, we vary $h$ over a candidate set $\Xi_h$
while fixing all other hyperparameters to their default values (marked by the dashed vertical line in the sweep plots).
For each setting, we run the full AutoQRA pipeline under the same evaluation budget and report the resulting final performance
(e.g., averaged task score; mean $\pm$ std over seeds).

To summarize sensitivity in a comparable scalar form, we define
\begin{equation}
\label{eq:hparam_sensitivity}
S(h)\triangleq \max_{\xi\in\Xi_h} P^\star(\xi)\;-\;\min_{\xi\in\Xi_h} P^\star(\xi),
\end{equation}
where $P^\star(\xi)$ denotes the best measured final score returned by AutoQRA under setting $h=\xi$.
Figure~\ref{fig:hparam_rank} ranks hyperparameters by $S(h)$, while Figure~\ref{fig:hparam_sweeps}
shows the full response curves for each sweep.

\noindent\textbf{Takeaway.}
Across the tested ranges, AutoQRA is generally stable around the default settings.
We use the ranking (Fig.~\ref{fig:hparam_rank}) to highlight which knobs matter most in practice,
and use the sweep curves (Fig.~\ref{fig:hparam_sweeps}) to make the directionality of each effect explicit.

\end{document}